\documentclass[10pt,twocolumn,letterpaper]{article}

\usepackage[pagenumbers]{cvpr} 

\usepackage[utf8]{inputenc}
\usepackage{caption, subcaption}  
\usepackage{multirow}

\usepackage{microtype}
\usepackage{xspace}
\usepackage{inconsolata}

\usepackage{graphicx}

\usepackage{tikz}
\usepackage{tikz-dependency}

\usepackage{todonotes}
\newcommand{\tvlm}{TE-VLM\xspace}
\newcommand{\tvlms}{TE-VLMs\xspace}

\definecolor{cvprblue}{rgb}{0.21,0.49,0.74}
\usepackage[pagebackref,breaklinks,colorlinks,allcolors=cvprblue]{hyperref}

\def\paperID{----} 
\def\confName{arXiv}
\def\confYear{2025}

\title{Seeing Syntax: Uncovering Syntactic Learning Limitations \\ in Vision-Language Models}

\author{Sri Harsha Dumpala$^{1*}$, David Arps$^{2*}$, Sageev Oore$^1$, Laura Kallmeyer$^2$, Hassan Sajjad$^1$ \\
$^1$Dalhousie University, Canada, $^2$Heinrich Heine University D\"usseldorf, Germany
}

\begin{document}
\maketitle
\begin{abstract}
Vision-language models (VLMs), serve as foundation models for multi-modal applications such as image captioning and text-to-image generation. Recent studies have highlighted limitations in VLM text encoders, particularly in areas like compositionality and semantic understanding, though the underlying reasons for these limitations remain unclear. In this work, we aim to address this gap by analyzing the syntactic information, one of the fundamental linguistic properties, encoded by the text encoders of VLMs. We perform a thorough analysis comparing VLMs with different objective functions, parameter size and training data size, and with uni-modal language models (ULMs) in their ability to encode syntactic knowledge. Our findings suggest that ULM text encoders acquire syntactic information more effectively than those in VLMs.
The syntactic information learned by VLM text encoders is shaped primarily by the pre-training objective, which plays a more crucial role than other factors such as model architecture, model size, or the volume of pre-training data. Models exhibit different layer-wise trends where CLIP performance dropped across layers while for other models, middle layers are rich in encoding syntactic knowledge.
\end{abstract}

\section{Introduction}

Vision-Language Models (VLMs), such as CLIP, have demonstrated remarkable performance across a range of downstream tasks, including image-text retrieval, image captioning, and text-to-image generation \cite{radford2021learning, li2023blip2, LiangWDLZ0ZVM23, dalle-2, imagen}. However, it remains unclear which linguistic features are encoded by the text encoders of VLMs to achieve this level of performance. Recent studies have demonstrated that Unimodal Language Models (ULMs), such as BERT~\cite{bert} and RoBERTa~\cite{liu2019roberta}, encode linguistic information, including semantic and syntactic sentence structure \cite{kulmizev-nivre-2022-schrodingers,mahowald-etal-2024-dissociating}. Do the text encoders of VLMs (\tvlm) also exhibit this ability to encode linguistic information, as ULMs do? 

In this direction, recent works have focused on analyzing the compositional abilities of \tvlm. These works have shown that 
\tvlms struggle to represent (encode) the word order in the input text; instead, 
they tend to represent the input text in a bag-of-words format~\cite{thrush2022winoground, zhao2022vl, ray2023cola, yuksekgonul2023and}. 
Sentence structure is important to correctly understand the meaning conveyed in an input sequence and is crucial for tasks such as image-text matching or image generation. 
Figure~\ref{fig:intro-fig} illustrates that even recent large VLMs often
lack the understanding of syntactic relations in a sentence. 
The sentence structure of the prompt, \textit{``A cat chases a dog"}, contains a clear description of the roles of the cat and the dog. 
This is expressed through the grammatical structure of the sentence in general, and through the grammatical roles \textit{nsubj} (nominal subject) and \textit{obj} (direct object) of the animals in particular. 
However, StableDiffusion 3.5 fails to consistently generate images in which the cat is the chaser (the \textit{nsubj} of the sentence). 
This means that either the text encoder does not appropriately encode the prompt structure, or the information is encoded but the image decoder fails to use it. 
In this study, we investigate the first of these possible causes.

\begin{figure}
    \begin{center}
    \begin{tabular}{rl}
    \includegraphics[width=.45\columnwidth]{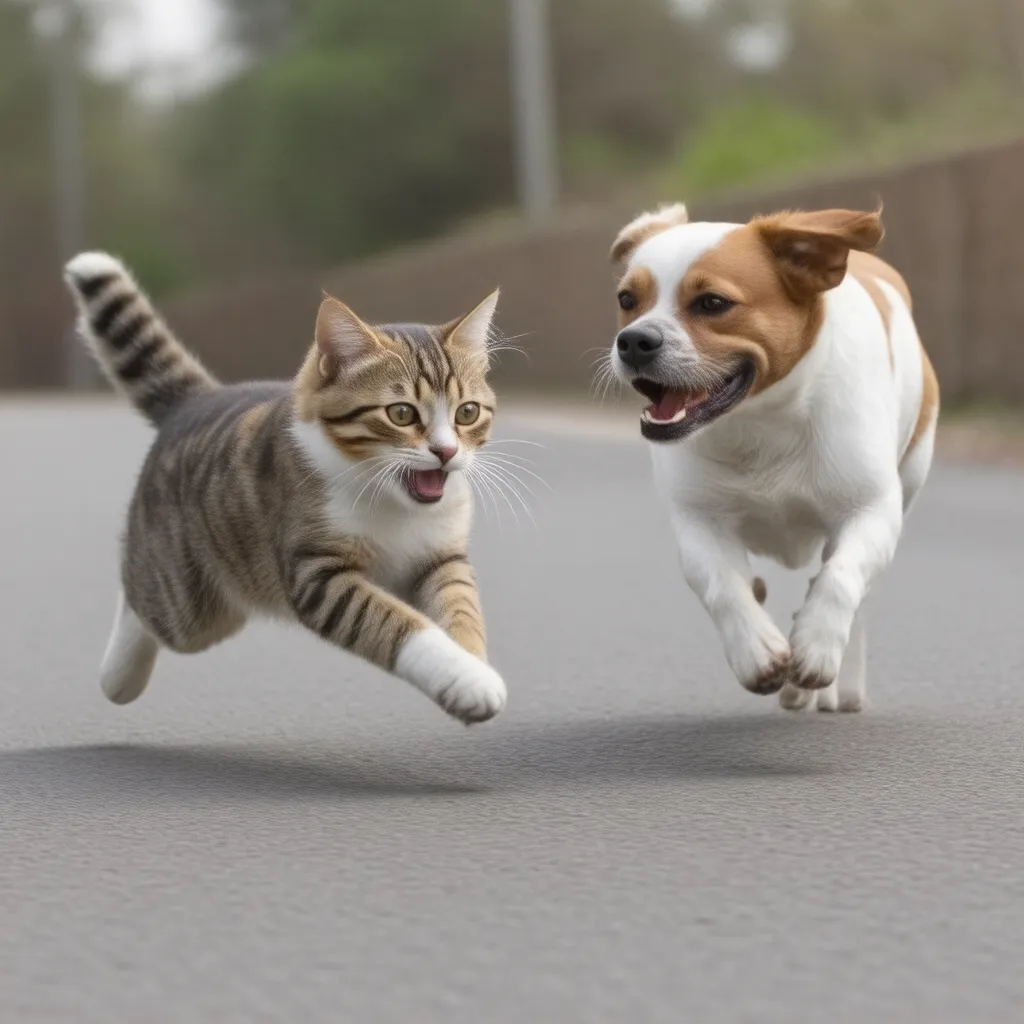} &
    \includegraphics[width=.45\columnwidth]{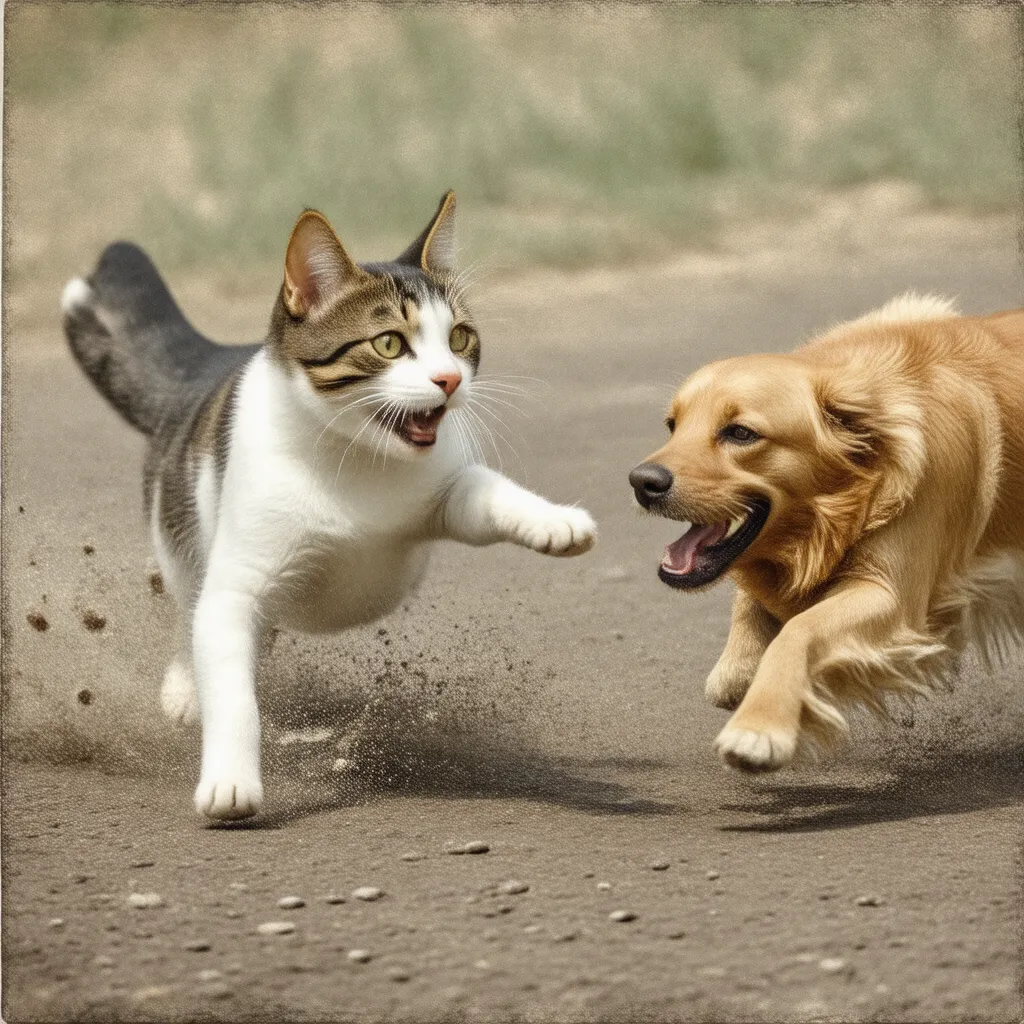} \\
    \includegraphics[width=.45\columnwidth]{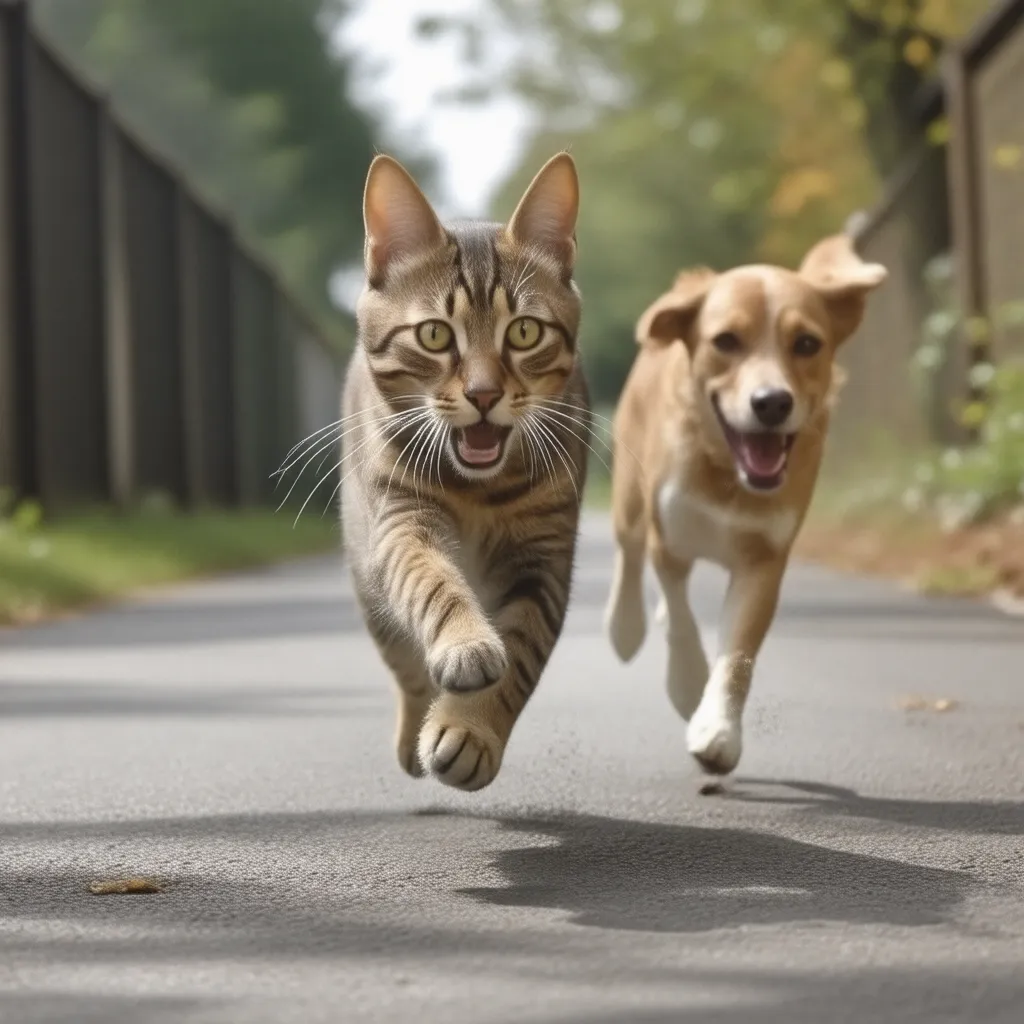} & 
    \includegraphics[width=.45\columnwidth]{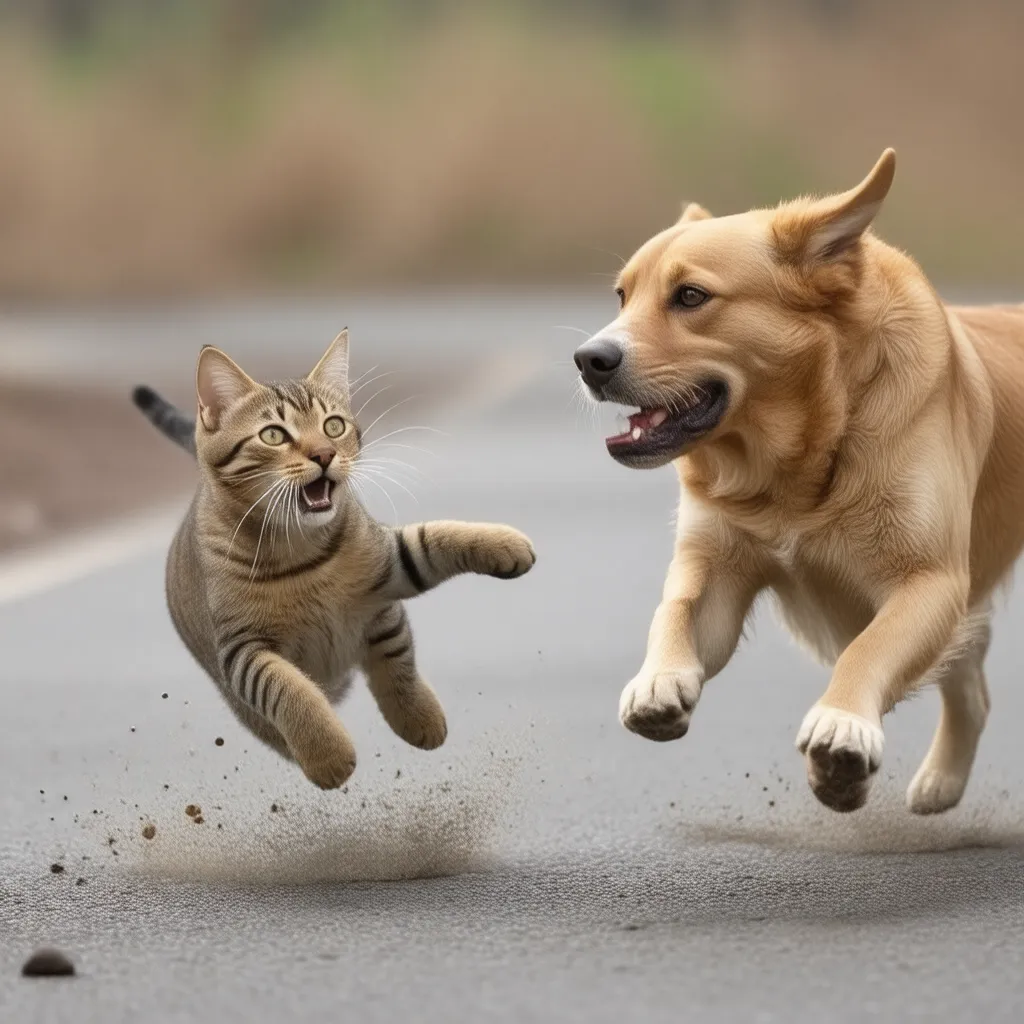} \\~\\
    \end{tabular}
    \vspace{1em}
    \begin{dependency}
        \begin{deptext}
        A \& cat \& chases \& a \& dog \\
        \end{deptext}
        \depedge{2}{1}{det}
        \depedge{3}{2}{nsubj}
        \deproot{3}{root}
        \depedge{5}{4}{det}
        \depedge{3}{5}{dobj}
    \end{dependency}
    \end{center}
    \vspace{-2em}
    \caption{Four images sampled from the StableDiffusion 3.5 model for the same prompt \textit{A cat chases a dog}, and the Universal Dependency tree for the prompt. StableDiffusion 3.5 uses CLIP-L, CLIP-bigG and T5-XXl models to encode the input prompt. 
    Accessed via \url{https://huggingface.co/spaces/stabilityai/stable-diffusion-3.5-large-turbo}.
    We experimented with variations of this prompt and various other prompts, with similar issues observed in the results.}
    \label{fig:intro-fig}
\end{figure}


While previous works have analyzed the distribution of linguistic information (both syntactic and semantic) across 
layers of ULMs \cite{kelly2020sentence, hewitt2019structural, kulmizev2020neural, li2021bert,jawahar2019does}, no studies have comprehensively evaluated the presence of and the distribution of syntactic information 
in
TE-VLMs. 
It is not yet clear how syntactic information is distributed across the layers of TE-VLMs or whether TE-VLMs exhibit similar patterns to ULMs. 
%
A major difference between VLMs and ULMs stems from the objective function used to pre-train these models. Most ULMs \cite{devlin-etal-2019-bert, liu2019roberta} are pre-trained using token-level objectives, i.e., predicting masked tokens, or predicting next tokens. Most VLMs, however, are trained using contrastive learning scheme to embed both images and text into a shared representation space. In this study, we analyze the distribution of syntactic information across the network of TE-VLMs and compare it with that of ULMs. 
To understand the impact of the pre-training objective function on linguistic information encoded by the VLMs, we analyze 
a variety of VLMs pre-trained with different objective functions in addition to contrastive loss objective function. 

To this end, our study aims to answer the main research question of ``Do VLMs encode syntactic knowledge?" We divide it into the following sub-questions:
\begin{enumerate}
\item How do the syntactic learning capabilities  of TE-VLMs compare to those of ULMs?
\item What is the influence of pre-training objective function in a VLM's ability to learn syntactic knowledge?
\item How does the syntactic information vary across the layers of TE-VLMs? 
\end{enumerate}


We use DepProbe~\cite{muller-eberstein-etal-2022-probing} to probe representations of VLMs in their ability to predict Universal Dependencies~\cite{de-marneffe-etal-2021-universal}. We perform experiments using various variants of CLIP~\cite{radford2021learning} and FLAVA~\cite{singh2022flava} and compare the results with several ULMs. Our notable findings are summarized below:

\begin{itemize}
    \item TE-VLMs perform poorly in learning syntax when compared to ULMs.
    \item Syntactic knowledge in representation is strongly influenced by the pre-training objective.
    \item Models exhibit different layer-wise trends where CLIP performance dropped across layers while other VLMs and ULMs show higher syntactic knowledge at middle layers.
    \item Using contrastive loss for training, increasing the number of parameters and the training data size do not improve the syntactic learning of a model. 
    \item CLIP struggles in learning relations between content words such as predicate-argument relations. However, functional relations such as determiners are predicted well by all models.
\end{itemize}







\section{Related work}

\subsection{Vision-language models (VLMs)}
Recent work has introduced several VLMs designed to learn the multimodal interactions between vision and language data. These models vary in architecture, size, pre-training objectives, and the amount of pre-training data \cite{radford2021learning, schuhmann2022laion, singh2022flava}. \citet{radford2021learning} introduced CLIP (Contrastive Language–Image Pre-training), which employs a contrastive learning framework to embed both images and text into a shared representation space. CLIP demonstrated impressive performance across several tasks including zero-shot image classification. Subsequent improvements to CLIP include increasing the model parameters and training dataset size \cite{schuhmann2022laion}, training with large number of noisy image-caption pairs \cite{jia2021scaling}, and incorporating additional objective functions beyond contrastive loss, such as masked language modeling, masked image modeling, and image-text matching \cite{singh2022flava, li2021albef, li2023blip2}.  The effects of these developments on the learning capabilities of VLMs, particularly regarding syntax understanding, have not been thoroughly explored in previous studies. In this work, we analyze the impact of these modifications on the syntactic understanding abilities of VLMs. 

\subsection{Sentence Structure in VLM text encoders}
Recent studies have focused on evaluating the linguistic and relational understanding abilities of VLMs~\cite{yuksekgonul2023and, wang2023can, kumar2024vision, toker-etal-2024-diffusion}. \citet{kumar2024vision} demonstrated the challenges VLMs face in understanding compound nouns by employing a text-to-image retrieval task, where the model must select the correct image depicting the compound noun in a given text prompt from a pair of distractor images. \citet{yuksekgonul2023and} highlighted the lack of compositional understanding in VLMs, presenting sentences such as \textit{the grass is eating the horse} to showcase model failures, despite the implausibility of such scenarios. \citet{toker-etal-2024-diffusion} analyzed the relationship between syntactic depth and noun order by generating images from the intermediate representations of text encoders of VLMs in a text-to-image pipeline using a diffusion model.

\citet{wang2023can} is the nearest neighbor to our work, as it analyzes the semantic and syntactic understanding capabilities of VLMs using an image-to-text retrieval task. While their study is similar to ours, it relies on image-based tasks to evaluate the models. In contrast, our approach focuses on analyzing the text encoders of VLMs and ULMs using only text as input, without requiring accompanying images. To allow for a more detailed analysis of linguistic knowledge, we focus on the relational information between text elements, represented through syntactic dependency trees. Additionally, we examine the distribution of syntactic information across the intermediate layers of VLMs and ULMs.

\subsection{Compare VLMs with ULMs}
\citet{alper2023bert} find that the CLIP text encoder outperforms ULMs in tasks requiring implicit visual reasoning, while \citet{ChenCDWW23} report that ULMs excel in general language understanding tasks as defined in the GLUE \cite{WangSMHLB19} benchmark. Moreover, recent studies show that VLMs struggle to encode compositionality in input text. More recent work \cite{dumpala2024sugarcrepe++, harsha2024sensitivity} demonstrates that VLM text encoders lack the ability to distinguish between lexical and semantic differences in text utterances. However, these studies do not provide sufficient explanations for these findings. In this study, we aim to differentiate the language understanding capabilities of VLM text encoders from those of ULMs. 
Syntactic structure is a crucial text property that encoders should capture during pre-training. Therefore, we analyze the ability of text encoders, both in VLMs and ULMs, to encode syntactic information from input text. Additionally, we examine how pre-training objectives, data size, and model architecture influence this ability in VLM text encoders. Finally, we compare the distribution of syntactic information across different layers of VLM text encoders with that of ULMs.

\subsection{Syntax Probing of ULMs}\label{sec:syntax-probing}
ULMs have been shown to encode linguistic properties such as information about morphology, syntax, semantics, etc. \cite{manning-etal-2020-emergent,linzen-baroni-2021-syntactic,kulmizev-nivre-2022-schrodingers,mahowald-etal-2024-dissociating}. 
These encodings are relatively robust to perturbations of word order, word co-occurences, and sentence meaning \cite{gulordava-etal-2018-colorless,sinha-etal-2021-masked,arps-etal-2022-probing,arps-etal-2024-multilingual}.
These works focus either on model behavior (i.e., predictions of a language modeling head) or representations (i.e. neural activations of the ULM when presented with a certain text). 
Because \tvlm{}s typically do not generate text, we focus on the latter methods here. 

These methods typically relate LM activations to linguistic representations such as the Universal Dependency (UD) tree depicted in Figure~\ref{fig:intro-fig}. 
Several approaches involve training diagnostic classifiers on frozen intermediate representations to predict Universal Dependencies (UD) trees \cite{hewitt-manning-2019-structural,kulmizev-etal-2020-neural,chi-etal-2020-finding,muller-eberstein-etal-2022-probing}. 
For instance, \cite{muller-eberstein-etal-2022-probing} probe multilingual BERT \cite{devlin-etal-2019-bert} in 13 languages and find that probing the middle layers yields the highest performance, that there are significant non-trivial relations between the probes trained on different languages, and that probing performance predicts the performance of a full parser after fine-tuning the LM.  


\section{Methodology}
Given a text encoder's representation of an input text, our goal is to probe whether certain syntactic relations are encoded in this representation. 
Concretely, we predict Universal Dependencies (UD, \cite{de-marneffe-etal-2021-universal}), a widely used annotation format for syntactic dependency trees. 
For instance, in the tree in Figure~\ref{fig:intro-fig}, \textit{dog} is a dependent of the verb \textit{chases} because there is a dependency edge from \textit{chases} to \textit{dog}.
The dependency edges are directed, where the direction indicates a relation from a head word to a dependent. 
Furthermore, each edge is labeled with a syntactic relation (specified in \cite{de-marneffe-etal-2021-universal}). For instance, \textit{dog} is the \textit{obj} (direct object) of \textit{chases}.
Several approaches exist for linearly separating UD trees (Sec.~\ref{sec:syntax-probing}. Of these, we use the approach from \citet{muller-eberstein-etal-2022-probing} because they are the only ones to decode both labeled and directed dependency trees. 

\subsection{Probing for Syntactic Structure}\label{sec:probing-method}
\paragraph{DepProbe} We use DepProbe \cite{muller-eberstein-etal-2022-probing}, a probing classifier that decodes syntactic dependency trees from word representations of text encoders. 
The model consists of two matrices, $L$ and $B$. 
Both matrices are implemented as linear neural network layers (without bias) that transform an LM representation vector of a word into a vector representing a syntactic property of that word. 
Assume that the LM representation of a word $w_i$ is $h_i \in \mathbb{R}^{d_h}$, and that the annotation has $l$ dependency relations. 
The matrix $L \in \mathbb{R}^{d_h \times l}$ is a linear classifier that predicts for each word the label $l$ of this word's incoming dependency edge -- the relation $r_i$ between the word and its head (\cite{muller-eberstein-etal-2022-probing}, Eq. 3):
\begin{align}
    p(r_i = l_k|w_i) = softmax(Lh_i)_k
\end{align}

This model component is trained using cross-entropy loss. 
The matrix $B
\in \mathbb{R}^{d_h \times b}$ predicts the dependency edges between words. 
$B$ projects the LM representations in a vector space that has less dimensions than the LM layer ($b < d_h$). 
$b$ is a hyperparameter, usually set to values up to 128. 
The target vector space is informally called the syntactic subspace: It reflects structural information such that vector distances between words mimic the distance between words in the dependency tree. 
Concretely, when $h_i, h_j$ are the LM representations of words $w_i, w_j$, their distance $d_{B}(h_i, h_j)$
in the syntactic subspace is defined by \citet[Eq.~1]{muller-eberstein-etal-2022-probing}:
\begin{align}
    d_{B}(h_i, h_j) = \sqrt{(Bh_i - Bh_j)^T(Bh_i - Bh_j)}
\end{align}

This model component is trained to predict the distance between all word pairs in the dependency tree. 
Assume that the distance between two words in the dependency tree is defined as the number of edges between the two words, $d_P(w_i, w_j)$. 
If $s$ is a sentence of length $N+1$, the loss for optimizing $B$ is given by \cite[Eq.~2]{muller-eberstein-etal-2022-probing}:
\begin{align}
    \mathcal{L}_{B}(s) = \frac{1}{N^2} \sum_{i=0}^{N} \sum_{j=0}^{N} | d_P(w_i, w_j) - d_{B}(h_i, h_j) |
\end{align}
The syntactic tree is rooted by $L$: The word with the highest probability of having the $root$ dependency label is the root of the dependency tree, and the rest of the tree is built in a top-down fashion. 
In cases where a word consists of multiple subword tokens, the token representations belonging to that word are aggregated by taking the element-wise mean. 

We report results for DepProbe in its default settings. We have conducted a hyperparameter search with no significant improvements over this setting. 

\paragraph{Evaluation}

\begin{figure}
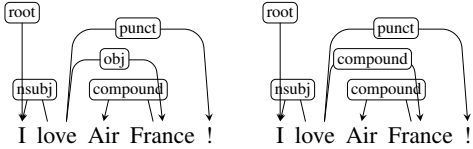

    \centering
    \scalebox{.9}{
    \begin{dependency}
    \begin{deptext}
    I \& love \& Air \& France \& ! \\
    \end{deptext}
    \depedge{2}{1}{nsubj}
    \deproot{1}{root}
    \depedge{4}{3}{compound}
    \depedge{2}{4}{obj}
    \depedge{2}{5}{punct}
    \end{dependency}
    \hspace{1em}
    \begin{dependency}
    \begin{deptext}
    I \& love \& Air \& France \& ! \\
    \end{deptext}
    \depedge{2}{1}{nsubj}
    \deproot{1}{root}
    \depedge{4}{3}{compound}
    \depedge{2}{4}{compound}
    \depedge{2}{5}{punct}
    \end{dependency}
}
    \caption{Example predictions. FLAVA and RoBERTa predict the gold tree (left), CLIP predicts the tree on the right}
    \label{fig:exampletree}
\end{figure}

We probe text encoders at different layers, and evaluate several aspects of the dependency tree using standard metrics from syntactic dependency parsing \cite[Ch.~19]{jurafsky-martin-2024-speech}: 
accuracy for dependency relation labels (LABEL); 
unlabeled attachment score (UAS) that counts the number of directed edges predicted correctly; 
UUAS is the undirected UAS where both the direction and labeling of an edge are ignored;
and labeled attachment score (LAS) that describes how many edges are predicted correctly with the right label.
Finally, ROOT indicates for how many sentences in the test set the root of the dependency tree is correctly identified. 
Consider the pair of example trees in Fig.~\ref{fig:exampletree}. 
The labeling accuracy of the tree predicted by CLIP (right) is $4/5=0.8$, because the the only incorrectly predicted label is the prediction of \texttt{compound} instead of \texttt{obj} for \textit{France}.
The UUAS is also 1, because all edges of the gold and predicted tree go in the same direction.
The unlabeled attachment score and the ROOT score are both one, because the predicted (unlabeled but directed) edges are the same, and the root is identified correctly. 
Therefore, the LAS is also 0.8. 
For readability, all scores are multiplied by 100 to be displayed as percentages. 



\subsection{Models}
\label{sec:mod_det}
In this paper, we examine three distinct categories of language models as explained below

\noindent \textbf{(1) Vision-Language Models (VLMs):} VLMs refer to multimodal self-supervised models trained using both visual and text modalities. We analyze different variants of CLIP and FLAVA.
\begin{itemize}
\item CLIP~\cite{radford2021learning}: a vision-language model pre-trained on a large dataset of image-caption pairs using an image-text contrastive (ITC) learning framework. In ITC learning, the model is trained to maximize the cosine similarity between the image and text embeddings of correct pairs while minimizing the cosine similarity for incorrect pairs. We analyze various CLIP model variants that differ in model size (ranging from 151 million to 1.2 billion parameters) and the volume of pre-training data (ranging from 400 million to 5 billion image-caption pairs). 
\item FLAVA~\cite{singh2022flava}: a VLM pretrained using ITC along with auxiliary objective functions such as , MLM, masked multimodal modeling (MMM), and masked image modeling (MIM). We analyze FLAVA model to understand the effect of these auxiliary objective functions on the syntactic learning abilities of VLMs.
\end{itemize}

\noindent \textbf{(2) Uni-modal Language Models (ULMs):} ULMs refer to self-supervised models pre-trained exclusively on text. Their pre-training typically involves the masked language modeling (MLM) task, where randomly selected words in the text are masked, and the model is trained to predict these masked words using the context provided by the surrounding unmasked words. 
In this paper, We use the following two different ULMs for comparison with the syntactic learning capabilities of TE-VLMs.

\begin{itemize}
\item RoBERTa~\cite{liu2019roberta}: a transformer-based language model pre-trained with the MLM task. We analyze two variants of RoBERTa -- RoBERTa-base and RoBERTa-large.
\item MiniLM~\cite{wang2020minilm}: a distilled language model produced by distilling the self-attention module from the last transformer layer of the teacher model. We analyze MiniLM distilled with BERT as the teacher model. 
\end{itemize}

\noindent\textbf{(3) Sentence Language Models (SLMs)~\cite{sbert}:} a modification of ULMs, trained on the natural language inference (NLI) task to derive semantically meaningful sentence-level (or global) embeddings. In this work, we analyze the syntactic learning capabilities of SLMs (Sentence MiniLM and Sentence RoBERTa-large), and draw parallels between SLMs and CLIP-based VLMs, as both are trained to learn global text embeddings.

\begin{table*}[!htb]
\centering
\resizebox{0.86\textwidth}{!}{
\begin{tabular}{lccc|cccccc}
\toprule
 \multirow{2}{*}{Model} & \#Total & Embedding & Pre-training & \multirow{2}{*}{Layer} & \multirow{2}{*}{LAS} & \multirow{2}{*}{UAS} & \multirow{2}{*}{UUAS} & \multirow{2}{*}{LABEL} & \multirow{2}{*}{ROOT} \\
   & Parameters & Dimension & Objectives &  &  &  &  &  & \\
 \midrule
\multirow{3}{*}{RoBERTa-base} & \multirow{3}{*}{125M} & \multirow{3}{*}{768} & \multirow{3}{*}{MLM} & 0 & \textbf{34.47} & \textbf{46.21} & \textbf{55.06} & 66.06 & 64.07 \\
& & & & 6 & 64.43 & 71.27 & 75.25 & 85.98 & 88.51 \\
& & & & 12 & 53.72 & 60.23 & 65.70 & 84.00 & 85.50 \\
\midrule
\multirow{3}{*}{MiniLM} & \multirow{3}{*}{33M} & \multirow{3}{*}{384} & \multirow{3}{*}{Distilled MLM} & 0 & 34.38 & 46.10	& 54.65 & \textbf{66.19} & \textbf{64.19} \\
& & & & 6 & \textbf{68.03} & \textbf{75.09} & \textbf{79.12} & \textbf{86.35} & \textbf{89.56} \\
& & & & 12 & \textbf{60.20} & \textbf{67.97} & \textbf{72.48} & \textbf{84.15} & \textbf{86.86} \\
 \midrule
\multirow{3}{*}{CLIP} & \multirow{3}{*}{151M} & \multirow{3}{*}{512} & \multirow{3}{*}{ITC} & 0 & 33.00 & 44.06 & 53.94 & \underline{64.93} & \underline{60.75} \\
& & & & 6 & \underline{26.25} & \underline{33.49} & \underline{38.91} & \underline{71.83} & \underline{72.60} \\
& & & & 12 & \underline{17.28} & \underline{24.66} & \underline{29.75} & \underline{59.30} & \underline{64.31} \\
\midrule
 \multirow{3}{*}{FLAVA} & \multirow{3}{*}{358M} & \multirow{3}{*}{768} & \multirow{3}{*}{ITC, MLM, MMM} & 0 & \underline{32.25} & \underline{43.70} & \underline{53.07} & 65.20 & 62.04 \\
& & & & 6 & 60.83 & 66.86 & 71.58 & 85.77 & 87.84 \\
& & & & 12 & 48.59 & 54.59 & 59.90 & 82.97 & 82.43 \\
 \bottomrule
\end{tabular}}
\caption{\small Compare performance of VLMs (CLIP and FLAVA) with ULMs (RoBERTa-base and Mini-LM) on the syntax probing task. Pre-training Objectives for the text encoders -- ITC: image-text contrastive; MLM: masked language modeling; MMM: masked multimodal modeling; \#Total parameters for CLIP and FLAVA include both image and text encoder models. Layer - 0, 6, 12 refers to the 0, 6 and 12 layers of the text encoders of the corresponding models. Best values are in bold, and lowest values are underlined for each layer and metric.} 
\label{tab:clip_rob}
\end{table*}

\subsection{Data}
We train and evaluate the probes on the EWT Universal Dependencies Treebank \citep{silveira-etal-2014-gold}. 
We filter out long sentences (above 75 tokens) to match the maximum sequence length of many text encoders.
The training split has a size of 10.8K sentences and 197K tokens. The development and test split each contain roughly 1.6K sentences, or 24K tokens. 
In the treebank, $l=37$ different syntactic relations are annotated. 
Furthermore, we evaluate the probes on the semantically perturbed Universal Dependencies (SPUD) data from \cite{arps-etal-2024-multilingual}, which contains nonsensical sentences with known syntactic structure. 
This tests to which extent probe results are indicative of syntax proper - i.e., purely grammatical information in the text encoder representation. 
This is achieved by testing on grammatical albeit nonsensical sentences with perturbed co-occurrence statistics for content words. 

\section{Results}

\begin{table*}[!htb]
\centering
\vspace{-0.3cm}
\resizebox{0.77\linewidth}{!}{
\begin{tabular}{lccc|cccccc}
\toprule
  \multirow{2}{*}{Model} & \#Total & Embedding & Pre-training & \multirow{2}{*}{Layer} & \multirow{2}{*}{LAS} & \multirow{2}{*}{UAS} & \multirow{2}{*}{UUAS} & \multirow{2}{*}{LABEL} & \multirow{2}{*}{ROOT} \\
   & Parameters & Dimension & data size &  &  &  &  &  & \\
 \midrule
 \multirow{3}{*}{CLIP-ViT-B/32} & \multirow{3}{*}{151M} & \multirow{3}{*}{512} & \multirow{3}{*}{400M} & 0 & 33.00 & 44.06 & 53.94 & 64.93 & 60.75 \\
 & & & & 6 & 26.25 & 33.49 & 38.91 & 71.83 & 72.60 \\
 & & & & 12 & 17.28 & 24.66 & 29.75 & 59.30 & 64.31 \\
 \midrule
\multirow{3}{*}{CLIP-ViT-B/16} & \multirow{3}{*}{151M} & \multirow{3}{*}{512} & \multirow{3}{*}{400M} & 0 & 32.86 & 43.92 & 53.65 & 65.03 & 61.21 \\
 & & & & 6 & 26.21 & 33.43 & 38.91 & 71.91 & 72.91 \\
 & & & & 12 & 17.28 & 24.66 & 29.75 & 59.30 & 64.31 \\
 \midrule
\multirow{3}{*}{LAION-CLIP-ViT-B/32} & \multirow{3}{*}{151M} & \multirow{3}{*}{512} & \multirow{3}{*}{2000M} & 0 & 31.99 & 43.37 & 53.46 & 64.24 & 59.77 \\
 & & & & 6 & 32.12 & 40.29 & 45.71 & 71.96 & 72.73 \\
 & & & & 12 & 17.50 & 26.09 & 31.02 & 58.20 & 66.09 \\
 \midrule
 \multirow{3}{*}{CLIP-XLM-RoB-Base-B/32} & \multirow{3}{*}{366M} & \multirow{3}{*}{512} & \multirow{3}{*}{5000M} & 0 & 22.56 & 32.84 & 43.32 & 46.98 & 57.06 \\
 & & & & 6 & 35.92 & 44.76 & 52.01 & 58.84 & 72.54 \\
 & & & & 12 & 15.68 & 22.72 & 28.71 & 50.88 & 63.33 \\
 \midrule
 \multirow{5}{*}{CLIP-XLM-RoB-Large-B/32} & \multirow{5}{*}{1193M} & \multirow{5}{*}{1024} & \multirow{5}{*}{5000M} & 0 & 22.24 & 32.65 & 42.73 & 46.93 & 56.70 \\
 & & & & 6 & 29.85 & 39.13 & 48.29 & 55.27 & 65.42 \\
 & & & & 12 & 39.56 & 48.28 & 54.96 & 60.30 & 72.50  \\
 & & & & 18 & 31.00 & 39.54 & 46.67 & 58.13 & 70.76 \\
 & & & & 24 & 14.58 & 20.73 & 26.14 & 52.34 & 64.50 \\
 \bottomrule
\end{tabular}}
\vspace{-0.1cm}
\caption{\small Performance of different variants of CLIP on the syntax probing task.  CLIP-ViT-B/32 is referred to as CLIP in this paper.} 
\label{tab:clip_var}
\vspace{-0.2cm}
\end{table*}

\begin{table}[!b]
\centering
\vspace{-0.3cm}
\resizebox{0.9\linewidth}{!}{
\begin{tabular}{llccccc}
\toprule
 & Layer & LAS & UAS & UUAS & LABEL & ROOT \\
 \midrule
 \multirow{3}{*}{CLIP-ViT-B/32} & 0 & 33.00 & 44.06 & 53.94 & 64.93 & 60.75 \\
 & 6 & 26.25 & 33.49 & 38.91 & 71.83 & 72.60 \\
 & 12 & 17.28 & 24.66 & 29.75 & 59.30 & 64.31 \\
 \midrule
\multirow{3}{*}{MiniLM} & 0 & \textbf{34.38} & \textbf{46.10} & \textbf{54.65} & \textbf{66.19} & \textbf{64.19} \\
& 6 & \textbf{68.03} & \textbf{75.09} & \textbf{79.12} & \textbf{86.35} & \textbf{89.56} \\
 & 12 & \textbf{60.20} & \textbf{67.97} & \textbf{72.48} & \textbf{84.15} & \textbf{86.86} \\
\midrule
\multirow{3}{*}{Sent-MiniLM} & 0 & 34.02 & 45.77 & 54.47 & 65.77 & 63.57 \\
 & 6 & 57.59 & 65.39 & 70.28 & 83.36 & 86.36 \\
 & 12 & 22.79 & 31.52 & 38.94 & 67.19 & 68.18 \\
 \midrule
 \midrule
 \multirow{4}{*}{RoBERTa-large} & 0 & \textbf{35.40} & 46.91 & 54.76 & \textbf{66.94} & \textbf{65.11} \\
& 6 & \textbf{63.62} & \textbf{70.01} & \textbf{74.05} & 85.94 & 86.43 \\
 & 12 & \textbf{62.37} & \textbf{69.34} & \textbf{73.40} & \textbf{85.80} & 87.47 \\
 & 24 & \textbf{50.02} & \textbf{56.18} & \textbf{60.66} & \textbf{83.80} & \textbf{85.69} \\
\midrule
\multirow{4}{*}{Sent-RoBERTa-large} & 0 & 35.36 & \textbf{47.13} & \textbf{55.09} & 66.72 & 64.93 \\
& 6 & 61.79 & 67.88 & 71.77 & \textbf{85.96} & \textbf{86.55} \\
 & 12 & 58.21 & 64.55 &69.03 & 85.54 & \textbf{88.64} \\
 & 24 & 19.98 & 28.37 & 35.42 & 60.21 & 69.35 \\
 \bottomrule
\end{tabular}}
\vspace{-0.1cm}
\caption{\small Compare performance of ULMs (MiniLM and RoBERTa-large) with the corresponding SLMs (Sentence-MiniLM and Sentence-RoBERTa-large). Best values in bold for each layer and each pair of ULM and Corresponding SLM. CLIP-ViT-B/32 performance values are provided here just for reference.}
\label{tab:lm_slm}
\end{table}

We evaluate the capabilities of the models listed in Section~\ref{sec:mod_det} to encode syntactic knowledge. It is to be noted that we evaluate only the text-encoder of VLMs (\tvlm), not the vision-encoder.
We summarize the main findings as follows: 

\paragraph{Pre-training objective influences encoding of syntactic information} Table~\ref{tab:clip_rob} presents the syntactic probing performance of ULMs (RoBERTa-base and MiniLM) and VLMs (CLIP and FLAVA). The results indicate that ULMs generally perform better in syntactic probing than TE-VLMs, especially when compared to CLIP. CLIP, trained with a contrastive (ITC) learning framework, exhibits lower syntactic encoding capabilities than ULMs at both the middle layer (layer 6) and final layer (layer 12). In contrast, FLAVA, another VLM that incorporates an MLM objective in addition to the contrastive loss, achieves performance closer to ULMs across all layers and metrics. Notably, MiniLM, a distilled model, performs similarly to RoBERTa-base. This shows the impact of pre-training objectives on the models ability to encode syntactic information, with the MLM objective encoding syntactic information more effectively than the ITC objective.

\paragraph{Do model size and pre-training data volume influence the syntactic learning capabilities of CLIP?}
To further support above observation, we evaluated a range of CLIP variants, all trained with a contrastive learning framework but differing in model size and pre-training data volume (Table~\ref{tab:clip_var}). We observed that the model and pre-training data size did not impact the syntactic learning capabilities of CLIP.
We elaborate the results as follows: 
\begin{enumerate}
\item \textit{Changing the patch size of CLIP's vision encoder does not affect the syntactic learning capabilities of the \tvlm}. We observe negligible to no difference in performance between CLIP-ViT-B/32 and CLIP-ViT-B/16, which are identical models except for patch sizes of 32 and 16, respectively.
\item \textit{Increasing the model size (total number of parameters) does not enhance the syntactic learning capabilities of CLIP.} The CLIP-ViT-B/32, CLIP-XLM-RoB-Base-B/32, and CLIP-XLM-RoB-Large-B/32 models achieve similar performance despite differing significantly in model size, with 151M, 366M, and 1,193M parameters, respectively. Notably, the text encoder of CLIP-XLM-RoB-Large-B/32 model consists of 24 encoder layers.
\item \textit{The volume of pre-training data does not affect the syntactic learning capabilities of CLIP}. LAION-CLIP-ViT-B/32, trained with 2 billion image-text pairs from LAION dataset, does not show any improvement in performance over CLIP-ViT-B/32, the same model trained with just 400 million image-text pairs. Similarly, CLIP-XLM-RoB-Base-B/32 and CLIP-XLM-RoB-Large-B/32 do not show any improvement even after being pre-trained with 5 billion image-text pairs. This suggests that increasing the pre-training data size may not enhance the syntactic learning capabilities of CLIP-based models.
\end{enumerate}


\paragraph{How do sentence language models encode syntactic information} We compare SLMs with ULMs to determine if SLMs exhibit behavior analogous to that of TE-VLMs. Table \ref{tab:lm_slm} compares ULMs and SLMs in their syntactic encoding capabilities. We observe an interesting pattern: in the initial layer (layer 0) and mid-layers (layer 6 for Sent-MiniLM and layer 12 for Sent-RoBERTa-large), SLMs behave similarly to ULMs, with only a slight degradation in performance. However, in the final layer (layer 12 or layer 24), the syntactic probing performance of SLMs is closer to that of CLIP. This behavior is intriguing, as the training objectives for both SLMs and CLIP models enable them to learn sentence-level embeddings without focusing heavily on finer details such as the syntactic structure of the text input. 

\paragraph{Variation across layers}
For all \tvlm{}s and ULMs, except the original CLIP models, we observe a general trend where the mid-layers have the highest probing performance (Tables \ref{tab:clip_rob}, \ref{tab:clip_var}). For most models and metrics, the performance of the final layer (usually layer 12) is higher than that of the initial layer (layer 0). Exceptions are listed in the next paragraph.
For CLIP, probing performance for most metrics deteriorates across layers, with an initial spike in lower layers (Figure~\ref{fig:layer_probes}). The comparable performance at layer 0 indicates that the (context-insensitive) word embeddings of all models initially contain similar levels of syntactic knowledge. In the middle layers, RoBERTa and FLAVA improve their encoding of syntactic knowledge, in contrast to CLIP, which shows a reduction in encoded knowledge compared to layer 0.

\begin{figure*}[t]
      \includegraphics[width=0.95\linewidth]{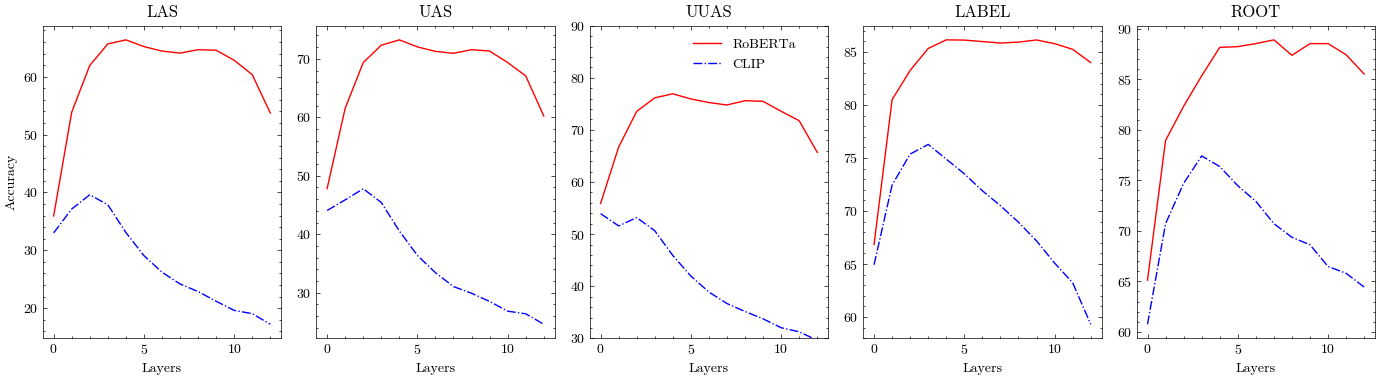}
      \caption{Comparison of the layer-wise performance between RoBERTa and CLIP. }
      \label{fig:layer_probes}
  \end{figure*}


\paragraph{Comparing metrics}
When comparing 
models, the trends reported above generally hold for all metrics. 
For all models, LABELing accuracy is higher than UAS - put differently, labeling the syntactic function of a word is easier than    determining its attachment place in the dependency tree.
Mostly, the layerwise trends reported above also hold across models.
Diverging from the layerwise trends reported above is that for Sent-MiniLM attachment scores which are higher at the embedding layer than at the final layer, but labeling score and the rooting score are higher at the final layer than at the embedding layer. 
This inversion is also visible for the rooting score (and sometimes labeling score) in CLIP variants. 
UUAS is higher than UAS by 4-9 points across models, indicating that recognizing correct edge directions bears the same level of difficulty for each model.

\begin{table}
    \centering
\resizebox{0.8\linewidth}{!}{
    \begin{tabular}{l|rrrrr}
        \toprule
        &\multicolumn{2}{c}{CLIP} & \multicolumn{2}{c}{Flava} & RoBERTa \\
        & score & $\Delta$ & score & $\Delta$ & score \\
        \midrule
        det  & 18.1 & 65.9 & 79.3 & 4.7 & 84.0 \\
        cc  & 22.0 & 53.7 & 63.4 & 12.3 & 75.7 \\
        case  & 10.6 & 50.0 & 60.0 & 0.6 & 60.6 \\
        mark  & 21.4 & 46.7 & 72.1 & -4.0 & 68.1 \\
        obj  & 44.1 & 45.2 & 85.5 & 3.8 & 89.3 \\
        nmod  & 26.1 & 41.6 & 64.1 & 3.6 & 67.7 \\
        amod  & 40.8 & 38.7 & 74.0 & 5.5 & 79.5 \\
        nsubj  & 40.4 & 38.6 & 71.5 & 7.5 & 79.0 \\
        compound  & 37.4 & 37.5 & 66.4 & 8.5 & 74.9 \\
        obl  & 26.0 & 37.3 & 57.1 & 6.2 & 63.3 \\
        cop  & 35.2 & 34.7 & 69.8 & 0.1 & 69.9 \\
        advmod  & 45.5 & 31.1 & 72.6 & 4.0 & 76.6 \\
        aux  & 51.4 & 27.4 & 82.8 & -4.0 & 78.8 \\
        conj  & 26.5 & 25.8 & 40.8 & 11.5 & 52.3 \\
        root  & 72.6 & 15.9 & 87.8 & 0.7 & 88.5 \\
        \bottomrule
        \end{tabular}
}
\caption{Attachment score for the 15 most common dependency relations in the ground truth of the UD test split, for the sixth layer of each model. CLIP refers to the ViT-B-P32 version. $\Delta$ shows the performance difference to RoBERTa (base).
}
\label{tab:results-by-deplabel}
\end{table}

\begin{table}\centering
    \resizebox{0.8\linewidth}{!}{
    \begin{tabular}{lrrrrr}
        \toprule
        &\multicolumn{2}{c}{CLIP} & \multicolumn{2}{c}{Flava} & RoBERTa \\
        & score & $\Delta$ & score & $\Delta$ & score \\
        \midrule
        compound & 42.4 & 38.8 & 80.0 & 1.2 & 81.2 \\
        obj & 55.6 & 28.0 & 84.5 & -0.9 & 83.6 \\
        aux & 72.4 & 24.6 & 97.2 & -0.1 & 97.1 \\
        obl & 45.8 & 21.2 & 68.2 & -1.2 & 67.0 \\
        mark & 75.2 & 21.1 & 96.4 & 0.0 & 96.4 \\
        conj & 61.2 & 19.2 & 77.8 & 2.5 & 80.4 \\
        nsubj & 74.5 & 18.9 & 92.5 & 0.9 & 93.4 \\
        nmod & 50.7 & 16.9 & 64.4 & 3.2 & 67.6 \\
        cop & 78.3 & 16.4 & 94.3 & 0.4 & 94.7 \\
        root & 72.6 & 15.9 & 87.8 & 0.7 & 88.5 \\
        amod & 76.5 & 14.7 & 91.2 & 0.1 & 91.2 \\
        advmod & 77.5 & 13.5 & 92.0 & -1.1 & 90.9 \\
        case & 89.7 & 7.1 & 97.2 & -0.3 & 96.9 \\
        det & 95.1 & 4.0 & 99.5 & -0.4 & 99.1 \\
        cc & 95.8 & 1.3 & 98.2 & -1.1 & 97.2 \\
        \bottomrule
        \end{tabular}}
    \caption{Labeling accuracy for the 15 most common relations in the ground truth of the UD test split, for the sixth layer of each model. $\Delta$ shows the performance difference to RoBERTa (base).}
    \label{tab:label-acc-by-deplabel}
\end{table}

\begin{figure}    \centering
    Gold
    \includegraphics[page=1,width=\linewidth, clip, trim=0.0cm 0.5cm 0.0cm 0.0cm]{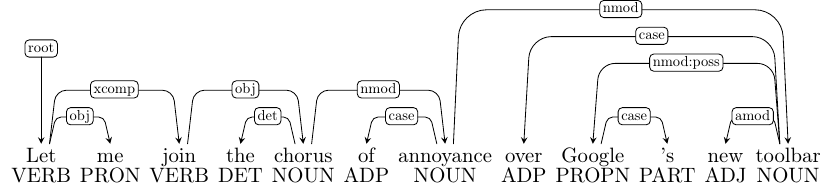}
    \vspace{.8em}\\
    CLIP
    \includegraphics[page=2,width=\linewidth, clip, trim=0.0cm 0.5cm 0.0cm 0.0cm]{fig/extreepreds1.pdf}
    \vspace{.8em}\\
    FLAVA
    \includegraphics[page=3,width=\linewidth, clip, trim=0.0cm 0.5cm 0.0cm 0.0cm]{fig/extreepreds1.pdf}
    \vspace{.8em}\\
    RoBERTa
    \includegraphics[page=4,width=\linewidth, clip, trim=0.0cm 0.5cm 0.0cm 0.1cm]{fig/extreepreds1.pdf}
    \caption{Example predictions with different types of errors.}
    \label{fig:exampletree2}
\end{figure}

\section{Discussion}
\paragraph{Investigating the Poor Performance of CLIP}
To investigate the reason for this poor performance, we observe the attachment score for some dependency relations in Table~\ref{tab:results-by-deplabel}. 
Across dependency relations, CLIP scores lowest, usually by huge margins to FLAVA and RoBERTa. 
Figures \ref{fig:exampletree} and \ref{fig:exampletree2} provides an illustration of these errors. 
In both examples, the handling of basic predicate-argument structure as well as function words (\textit{the, of, over}) is learned incorrectly, as outlined below. 
For basic predicate-argument structure, the core relations \texttt{nsubj} (nominal subject) or \texttt{obj}ect are frequently used to mark the most important entities in a sentence. 
In Figure~\ref{fig:exampletree}, the probe of CLIP predicts the relation and the attachment of the subject \textit{I} correctly. 
The object \textit{Air France} is built correctly as a compound, but is attached incorrectly to the main verb (using the relation label \texttt{compound} instead of \texttt{obj}). 
Similarly, in Figure~\ref{fig:exampletree2}, the probe of CLIP incorrectly predicts that \textit{chorus} is the \texttt{nsubj} of \textit{Let} (whereas it is annotated to be the object of \textit{join}). 
Furthermore, the corresponding object phrase \textit{the chorus of annoyance over Google's new toolbar} is built incorrectly. This phrase is the object of \textit{join} because all words in that phrase (recursively) depend on \textit{chorus}, i.e. \textit{chorus} is the head of that phrase. 
This phrase is built primarily through modification relations (\texttt{nmod} is nominal modification, \texttt{nmod:poss} describes that \textit{Google} \texttt{poss}esses the \textit{toolbar}). 
The probe for CLIP instead predicts that \textit{annoyance over Google's new toolbar} is a nominal modifier of \textit{let}, that the nouns in that phrase stand in a \texttt{compound} relation, and that \textit{new} is an \texttt{amod} (adjectival modifier) of \textit{'s} instead of \textit{toolbar}. 
The probes for FLAVA and RoBERTa predict correct trees for Figure~\ref{fig:exampletree}.

On the other hand, CLIP also scores much lower than the other models in functional relations such as attaching \texttt{case} markers, \texttt{det}erminers, or conjunctions (\texttt{cc}). In these cases, inspecting the results shows a tendency for the CLIP probe to attach these words in a way that is not linguistically meaningful. For instance, function words are attached to other function words instead of their content word head, presumably because the relation between a function word and its head is not linearly separable from the representations. 
Similarly, in the trees predicted by the CLIP probe, there is a tendency that the attachment violates simple rules of word order. 
While the annotated training data of the probe assumes that \textit{pre}positions are attached to a following noun, CLIP often attaches prepositions to preceding nouns (\textit{of} and \textit{over} in Figure~\ref{fig:exampletree2}).
These examples are complemented by the labeling accuracy for different dependency relations (displayed in detail in Table~\ref{tab:label-acc-by-deplabel}).
These show that functional relations (such as determiners, conjunctions, case markers) are predicted almost as well as the probe for CLIP than by the probes for RoBERTa. 
For relations between content words (different kinds of \texttt{mod}ifiers, predicate-argument relations, etc.), the probe on Flava performs closest to the probe on RoBERTa while scores on CLIP are far worse.



\noindent \textbf{Does DepProbe generalize?} 
DepProbe trains additional parameters on frozen text encoder representations (Sec.~\ref{sec:probing-method}).
This brings the question whether DepProbe generalizes beyond the training data: Does DepProbe on CLIP yield poor performance on held-out data at test time, or does DepProbe fail to generalize from the probe training data to probe testing data? To answer this question, we compare train and test set performance for CLIP in Table~\ref{tab:train-results}. 
The difference between both datasets is relatively small, showing that DepProbe does not capture the probe training data for CLIP. 

\begin{table}\centering
    \resizebox{0.4\textwidth}{!}{
    \begin{tabular}{ll|ccccc}
    \toprule
    Layer & split & LAS & UAS & UUAS & LABEL & ROOT\\
    \midrule
    \multirow{2}{*}{0} & train & 31.87 & 43.72 & 53.96 & 64.94 & 58.16 \\
    &                     test & 33.00 & 44.06 & 53.94 & 64.93 & 60.75 \\
    \midrule
    \multirow{2}{*}{6} & train & 26.43 & 33.86 & 39.10 & 73.15 & 75.58 \\
    &                     test & 26.25 & 33.49 & 38.91 & 71.83 & 72.60 \\
    \midrule
    \multirow{2}{*}{12} & train & 16.77 & 24.37 & 29.76 & 60.71 & 62.12 \\
    &                      test & 17.28 & 24.66 & 29.75 & 59.30 & 64.31 \\
    \bottomrule
    \end{tabular}}
    \caption{Comparing DepProbe results for CLIP on train and test splits of the probe training data.}
    \label{tab:train-results}
\end{table}

\begin{table}[!tb]
\label{tab:clip_rob_nonce}
\centering
\resizebox{0.95\linewidth}{!}{
\begin{tabular}{llccccc}
\toprule
 & Layer & LAS & UAS & UUAS & LABEL & ROOT \\
 \midrule
\multirow{3}{*}{RoBERTa-base} & 0 & 32.77 & 44.30 & 52.87 & 65.25 & 62.97 \\
 & 6 & 60.11 & 67.63 & 72.07 & 83.34 & 84.46 \\
 & 12 & 48.75 & 55.60 & 61.55 & 80.52 & 80.97 \\
 \midrule
 \multirow{3}{*}{Sent-MiniLM} & 0 & 31.25 & 42.76 & 51.60 & 64.38 & 62.55 \\
 & 6 & 55.44 & 63.15 & 68.95 & 81.39 & 80.97 \\
 & 12 & 22.08 & 30.04 & 37.62 & 67.56 & 66.52 \\
 \midrule
 \multirow{3}{*}{CLIP} & 0 & 29.17 & 40.15 & 50.28 & 60.83 & 59.12 \\
 & 6 & 23.02 & 30.29 & 35.89 & 66.37 & 68.97 \\
 & 12 & 14.82 & 22.24 & 27.68 & 54.53 & 58.63 \\
 \midrule
 \multirow{3}{*}{FLAVA}& 0 & 29.54 & 41.01 & 50.49 & 63.77 & 61.14 \\
 & 6 & 56.84 & 63.22 & 68.69 & 83.42 & 83.97 \\
 & 12 & 44.05 & 50.13 & 55.78 & 80.86 & 80.23 \\
 \bottomrule
\end{tabular}}
\caption{Compare performance of VLMs (CLIP and FLAVA) with an ULM (RoBERTa-base) and SLM (Sentence-MiniLM) on nonsensical sentences.}
\label{tab:nonce_res}
\end{table}

\noindent \textbf{Probing on nonsensical data} 
Furthermore, we probe the text encoders on nonsensical data with known syntactic structure, in order to separate out effects of lexical co-occurence from syntactic function. For this, we test on data obtained from \cite{arps-etal-2024-multilingual}. The results (in Table~\ref{tab:nonce_res}) consistently show 
slightly lower scores per model than on original data (Tables \ref{tab:clip_rob}, \ref{tab:lm_slm}).
This small but consistent performance drop on nonsensical data is analoguous to what has been found for ULMs by \cite{arps-etal-2024-multilingual}. 
This suggests that the syntactic information in the representations of each model is to a large extent independent from semantic (word co-occurence) information.  




\section{Conclusion}\label{sec:conclusion}
We probed the text encoders of VLMs (TE-VLMs) for their ability to encode syntactic knowledge. Our analysis showed that TE-VLMs are relatively poor at learning syntax compared to uni-modal language models (ULMs). Moreover, the pre-training objective strongly influences how text representations capture syntactic information. In particular, token-based objectives such as masked language modeling encourage the linear separability of dependency tree properties from the representations. CLIP-style models, which are trained solely on a sentence-level contrastive loss function, fail to capture even basic features of the dependency tree, such as word order or predicate-argument relations (e.g., subject and object). Our findings serve as an intrinsic evaluation that highlights the limitations of models trained using contrastive loss alone and support the use of multiple auxiliary objective functions to better capture language properties.



{\small
\bibliographystyle{ieeenat_fullname}
\bibliography{references,anthology1,anthology2}
}












\newpage
\appendix

\section{Implementation} 

For instructions to run an example probing experiment, please 
%
%
%
%
%
%
%
%
%

\subsection{Probing}

The experiments were conducted using the public implementation of DepProbe (\url{https://github.com/personads/depprobe}).

\section{Details of the Models Evaluated in this Study}
Below we provide the details of the model evaluated in this study.
\begin{itemize}
\setlength\itemsep{1.1em}
\item \textbf{Vision-Language Model (VLMs)}
\begin{enumerate}
\setlength \itemsep{0.4em}
\item CLIP~\cite{radford2021learning}: 12-layered transformer encoder model trained on 400 million image-text pairs. We evaluated the 'ViT-B/32' variant of CLIP -- ViT base model trained with a image patch size of 32 -- publicly available at the following \href{https://huggingface.co/openai/clip-vit-base-patch32}{HuggingFace Link}
\item FLAVA~\cite{singh2022flava}: 12-layered transformer encoder model trained on 70 million image-text pairs. We evaluated the FLAVA pre-trained Model available at the following \href{https://huggingface.co/facebook/flava-full}{HuggingFace Link}
\end{enumerate}
\item \textbf{Unimodal Language Models (ULMs)}
\begin{enumerate}
\setlength \itemsep{0.4em}
\item RoBERTa-base~\cite{liu2019roberta}: 12-layered transformer network trained exclusively on text data. We evaluated the RoBERTa-base model available at the following \href{https://huggingface.co/FacebookAI/roberta-base}{HuggingFace Link}
\item RoBERTa-large~\cite{liu2019roberta}: 24-layered transformer network trained exclusively on text data. We evaluated the RoBERTa-large model available at the following \href{https://huggingface.co/FacebookAI/roberta-large}{HuggingFace Link}
\item MiniLM~\cite{wang2020minilm}: 12-layered transformer network trained exclusively on text data by distilling the BERT pre-trained model. We evaluated the MiniLM model available at the following \href{https://huggingface.co/microsoft/MiniLM-L12-H384-uncased}{HuggingFace Link}
\end{enumerate}
\item \textbf{Sentence Language Models (SLMs)}~\cite{sbert}
\begin{enumerate}
\setlength \itemsep{0.4em}
\item Sentence-MiniLM~\cite{sbert}: Sentence language model trained on top of MiniLM (explained above). We evaluated the sentence-MiniLM model available at the following \href{https://huggingface.co/sentence-transformers/all-MiniLM-L12-v2}{HuggingFace Link}
\item Sentence-RoBERTa-large~\cite{sbert}: Sentence language model trained on top of RoBERTa-large model (explained above). We evaluated the sentence-MiniLM model available at the following \href{https://huggingface.co/sentence-transformers/all-roberta-large-v1}{HuggingFace Link}
\end{enumerate}
\end{itemize}

\paragraph{Variants of CLIP}
Below, we provide details of the various CLIP model variants evaluated in this study (Results provided in Table 2 of the main paper). 

\begin{enumerate}
\setlength\itemsep{0.5em}
\item CLIP-ViT-B/32~\cite{radford2021learning}: ViT base model with 12 transformer layers as the backbone of the text encoder. Here the images are provided as input with a patch size of 32. This model is trained using the WebImageText dataset~\cite{radford2021learning} consisting of 400M image-text pairs. We evaluated the CLIP-ViT-B/32 model  publicly available at the following \href{https://huggingface.co/openai/clip-vit-base-patch32}{HuggingFace Link}.
\item CLIP-ViT-B/16~\cite{radford2021learning}: This model is same as the CLIP-ViT-B/32 model but here the images are provided as input with a patch size of 16. This model is also trained using the WebImageText dataset~\cite{radford2021learning} consisting of 400M image-text pairs. We evaluated the CLIP-ViT-B/16 model  publicly available at the following \href{https://huggingface.co/openai/clip-vit-base-patch16}{HuggingFace Link}.
\item LAION-CLIP-ViT-B/32~\cite{schuhmann2022laion}: This model is similar to the CLIP-ViT-B/32 model but this model is trained using the 2 billion image-text pairs from the LAION dataset~\cite{schuhmann2022laion}. We evaluated the LAION-CLIP-ViT-B/32 model publicly available at the following \href{https://huggingface.co/laion/CLIP-ViT-B-32-laion2B-s34B-b79K}{HuggingFace Link}.
\item CLIP-XLM-RoBERTa-base-ViT-B/32 (referred to as CLIP-XLM-RoB-Base-B/32 in Table 2 of main paper)~\cite{schuhmann2022laion}: The text encoder of this model is initialized with RoBERTa-base weights and subsequently pre-trained using a contrastive learning framework. The pre-training process utilized 5 billion image-text pairs from the LAION dataset. We evaluated the model publicly available at the following \href{https://huggingface.co/calpt/CLIP-ViT-B-32-xlm-roberta-base-laion5B-s13B-b90k}{HuggingFace Link}.
\item CLIP-XLM-RoBERTa-large-ViT-H/14 (referred to as CLIP-XLM-RoB-Large-B/32 in Table 2 of main paper)~\cite{schuhmann2022laion}: The text encoder of this model is initialized with RoBERTa-large weights and subsequently pre-trained using a contrastive learning framework. The pre-training process utilized 5 billion image-text pairs from the LAION dataset. We evaluated the model publicly available at the following \href{https://huggingface.co/calpt/CLIP-ViT-H-14-frozen-xlm-roberta-large-laion5B-s13B-b90k}{HuggingFace Link}.
\end{enumerate}

\section{Treebanks}

\begin{figure}
    \centering
    \vspace{0.3cm}
    \includegraphics[width=1.0\linewidth]{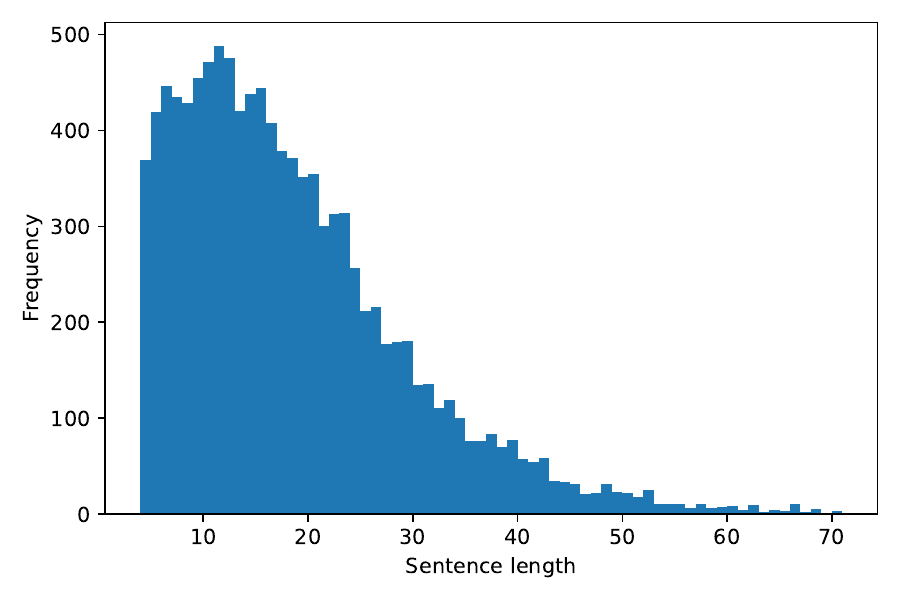}
    \caption{Sentence length distribution in thee filtered EWT training split.}
    \label{fig:sent-lengths-hist}
\end{figure}

We primarily use the EWT Universal Dependencies Treebank \cite{silveira-etal-2014-gold}. 
For evaluation on nonsensical sentences, we use the SPUD data. 
Examples from this dataset are available in \cite{arps-etal-2024-multilingual}.
After filtering out long sentences, the sentence lengths (in the filtered EWT training split) range from 4 to 71, with a mean (std.) of 18.1 (11.0). 
The distribution of sentence lengths is displayed in Figure \ref{fig:sent-lengths-hist}.

\section{Additional Experimental Results}

Table~\ref{tab:rob_clip_flava_layer} provides the layer-wise comparison between RoBERTa, CLIP and FLAVA in terms of their syntactic probing performance. We can make the following observations:

\begin{itemize}
\setlength\itemsep{0.7em}
\item RoBERTa (a ULM trained using masked language modeling (MLM) task) achieves the best performance among the three models across all layers. As we go deeper into the model, the syntactic probing performance improves up to Layer 6, followed by a slight degradation in performance toward the final layer (Layer-12). This indicates that the mid-layers encode the highest level of syntactic information.
\item CLIP models exhibit the lowest performance among the three models. In the initial layers, there is a slight improvement in performance (up to Layer 3), but the performance decreases as we go deeper into the model. The final layers achieve very low performance, signifying the difficulty of CLIP-based models in encoding syntactic information.
\item FLAVA, another VLM trained with contrastive loss and MLM as objective functions, shows behavior similar to RoBEERTa. Similar to RoBERTa, FLAVA shows an improvement in performance up to mid-layers (Layer 5), followed by a degradation in performance toward the final layer (Layer 12). This shows the importance of pre-training objectives in enhancing a model's ability to encode syntactic information.
\end{itemize}
The above observations signify the importance of including MLM as a pre-training objective for VLMs to enable encoding of syntactic information.

\begin{table*}[!htb]
\vspace{0.9em}
\centering
\resizebox{1\linewidth}{!}{
\begin{tabular}{l|ccccc|ccccc|ccccc}
\toprule
 \multicolumn{1}{c}{} & \multicolumn{5}{c}{RoBERTa} & \multicolumn{5}{c}{CLIP} & \multicolumn{5}{c}{FLAVA} \\
\cmidrule(lr){2-6} \cmidrule(lr){7-11} \cmidrule(lr){12-16}
 & LAS & UAS & UUAS & LAB & ROOT & LAS & UAS & UUAS & LAB & ROOT & LAS & UAS & UUAS & LAB & ROOT  \\
\midrule
Layer-0 & \textbf{35.93} & \textbf{47.81} & \textbf{55.88} & \textbf{66.84} & \textbf{65.11} & 33.00 & 44.06 & 53.94 & 64.93 & 60.75 & 32.25 & 43.70 & 53.07 & 65.20 & 62.04 \\
Layer-1 & \textbf{53.87} & \textbf{61.54} & \textbf{66.71} & \textbf{80.50} & \textbf{78.93} & 37.13 & 45.88 & 51.59 & 72.45 & 70.70 & 49.72 & 57.59 & 63.82 & 78.94 & 76.78 \\
Layer-2 & \textbf{61.95} & \textbf{69.35} & \textbf{73.55} & \textbf{83.26} & \textbf{82.31} & 39.59 & 47.78 & 53.18 & 75.37 & 74.69 & 51.51 & 59.32 & 65.07 & 79.75 & 77.89 \\
Layer-3 & \textbf{65.67} & \textbf{72.33} & \textbf{76.16} & \textbf{85.33} & \textbf{85.32} & 37.87 & 45.43 & 50.67 & 76.28 & 77.40 & 56.91 & 63.64 & 68.79 & 83.20 & 83.11 \\
Layer-4 & \textbf{66.37} & \textbf{73.24} & \textbf{76.93} & \textbf{86.14} & \textbf{88.14} & 33.07 & 40.56 & 45.92 & 74.89 & 76.35 & 59.06 & 65.83 & 70.97 & 83.51 & 83.11 \\
Layer-5 & \textbf{65.21} & \textbf{72.04} & \textbf{75.94} & \textbf{86.12} & \textbf{88.21} & 29.08 & 36.38 & 41.97 & 73.51 & 74.45 & 62.36 & 68.41 & 73.01 & 86.18 & 87.10 \\
Layer-6 & \textbf{64.43} & \textbf{71.27} & \textbf{75.25} & \textbf{85.98} & \textbf{88.51} & 26.21 & 33.43 & 38.91 & 71.91 & 72.91 & 60.83 & 66.86 & 71.58 & 85.77 & 87.84 \\
Layer-7 & \textbf{64.08} & \textbf{70.94} & \textbf{74.78} & \textbf{85.84} & \textbf{88.88} & 24.18 & 31.05 & 36.70 & 70.50 & 70.70 & 59.27 & 65.49 & 70.41 & 85.38 & 87.22 \\
Layer-8 & \textbf{64.67} & \textbf{71.56} & \textbf{75.61} & \textbf{85.93} & \textbf{87.35} & 22.86 & 29.91 & 35.17 & 68.94 & 69.35 & 58.92 & 65.04 & 70.03 & 85.73 & 87.84 \\
Layer-9 & \textbf{64.58} & \textbf{71.33} & \textbf{75.48} & \textbf{86.13} & \textbf{88.51} & 21.19 & 28.51 & 33.77 & 67.18 & 68.61 & 57.00 & 63.16 & 68.19 & 84.91 & 85.87 \\
Layer-10 & \textbf{62.84} & \textbf{69.37} & \textbf{73.58} & \textbf{85.78} & \textbf{88.51} & 19.61 & 26.85 & 32.06 & 65.09 & 66.46 & 54.36 & 60.61 & 65.81 & 84.15 & 84.15 \\
Layer-11 & \textbf{60.35} & \textbf{67.08} & \textbf{71.77} & \textbf{85.25} & \textbf{87.41} & 19.04 & 26.40 & 31.28 & 63.24 & 65.79 & 50.17 & 56.18 & 61.45 & 83.37 & 82.74 \\
Layer-12 & \textbf{53.72} & \textbf{60.23} & \textbf{65.70} & \textbf{84.00} & \textbf{85.50} & 17.21 & 24.61 & 29.65 & 59.32 & 64.43 & 48.59 & 54.59 & 59.90 & 82.97 & 82.43 \\
 \bottomrule
\end{tabular}}
\vspace{0.2cm}
\caption{Layer-wise syntax probing performance comparison between RoBERTa (a ULM trained using masked language modeling (MLM) task), CLIP (a VLM trained using only contrastive loss, and FLAVA (a VLM trained using multiple objectives, including contrastive loss and MLM). It can be observed that FLAVA's syntactic learning pattern is more similar to RoBERTa (the model with the best performance) than to CLIP. This highlights the importance of the pre-training objective in learning syntactic information. This table is an extension of \textcolor{magenta}{Table 1} in the main paper.}
\label{tab:rob_clip_flava_layer}
\end{table*}

\begin{table*}[!htb]
\centering
\resizebox{0.75\linewidth}{!}{
\begin{tabular}{l|ccccc|ccccc}
\toprule
 \multicolumn{1}{c}{} & \multicolumn{5}{c}{MiniLM} & \multicolumn{5}{c}{Sentence-MiniLM} \\
\cmidrule(lr){2-6} \cmidrule(lr){7-11}
 & LAS & UAS & UUAS & LAB & ROOT & LAS & UAS & UUAS & LAB & ROOT  \\
\midrule
Layer-0 & 34.38 & 46.10 & 54.65 & 66.19 & 64.19 & 34.02 & 45.77 & 54.47 & 65.77 & 63.57 \\
Layer-1 & 43.18 & 53.27 & 60.70 & 72.98 & 72.05 & 41.63 & 51.87 & 59.81 & 72.28 & 71.93 \\
Layer-2 & 56.45 & 64.57 & 69.77 & 80.59 & 80.34 & 54.12 & 62.52 & 68.22 & 79.37 & 78.87 \\
Layer-3 & 63.40 & 70.44 & 74.69 & 84.43 & 84.71 & 59.64 & 67.27 & 71.95 & 82.51 & 82.68\\
Layer-4 & 66.92 & 73.60 & 77.67 & 85.59 & 86.92 & 62.45 & 69.69 & 74.03 & 84.21 & 86.18\\
Layer-5 & 67.58 & 74.58 & 78.65 & 86.08 & 88.70 & 61.17 & 68.56 & 72.98 & 84.59 & 88.02 \\
Layer-6 & 68.03 & 75.09 & 79.12 & 86.35 & 89.56 & 57.59 & 65.39 & 70.28 & 83.36 & 86.36\\
Layer-7 & 66.27 & 73.68 & 77.92 & 85.88 & 90.23 & 51.76 & 59.55 & 65.51 & 82.57 & 86.18 \\
Layer-8 & 65.64 & 72.99 & 77.32 & 85.86 & 89.80 & 48.13 & 56.45 & 62.95 & 80.87 & 82.49 \\
Layer-9 & 63.62 & 70.64 & 74.91 & 85.75 & 89.80 & 37.18 & 45.29 & 51.69 & 78.50 & 79.85 \\
Layer-10 & 61.52 & 69.13 & 73.57 & 84.71 & 88.27 & 29.11 & 37.43 & 43.93 & 74.38 & 74.88 \\
Layer-11 & 58.87 & 66.13 & 70.53 & 84.41 & 87.16 & 22.36 & 30.32 & 37.35 & 69.85 & 70.27 \\
Layer-12 & 60.20 & 67.97 & 72.48 & 84.15 & 86.86 & 22.79 & 31.52 & 38.94 & 67.19 & 68.18 \\
 \bottomrule
\end{tabular}}
\vspace{0.1cm}
\caption{Layer-wise syntax probing performance comparison between MiniLM (a ULM) and Sentence-MiniLM (a SLM). It can be observed that Sentence-MiniLM follows a pattern similar to MiniLM up to the mid-layer (Layer 6). However, beyond Layer 6, the performance of Sentence-MiniLM drops drastically compared to MiniLM. This table extends \textcolor{magenta}{Table 3} in the main paper, where we compare the performance of ULMs with that of SLMs. }
\label{tab:mini_sent_mini_layer}
\end{table*}

Table \ref{tab:mini_sent_mini_layer} compares the layer-wise syntactic probing performance of MiniLM (a ULM) and Sentence-MiniLM (an SLM). MiniLM achieves performance comparable to RoBERTa across all layers. Sentence-MiniLM exhibits behavior similar to MiniLM up to the mid-layers (Layer 6). However, its performance diverges from MiniLM, with Sentence-MiniLM showing a drastic drop as we approach the final layers (Layer 12). \\


Figure \ref{fig:layer_probes} shows the layer-wise comparison between the three different models, RoBERTa (a ULM), CLIP (a VLM) and Sentence-MiniLM (a SLM). The main observations are as follows:

\begin{itemize}
\setlength \itemsep{0.5em}
\item \textbf{RoBERTa -} the syntax information encoded in RoBERTa increases up to mid-layers (Layer-6). Beyond Layer-6, there is a slight degradation in the performance as we go towards the final layers (Layer-12).
\item \textbf{CLIP -} Although there is a slight improvement in the syntactic information encoded in the first two or three layers, CLIP typically shows significant degradation in syntactic information as we move toward the higher layers.
\item \textbf{Sentence-MiniLM (Sent-MiniLM) -} shows an interesting pattern: in the initial layers, it follows RoBERTa, while from the mid-layers onwards, it exhibits a pattern similar to that of CLIP. This shows that the sentence-level training of SLM preserves syntactic information in the initial layers but leads to degradation in the syntactic information encoded in the mid-layers (Layer 6) to the final layers (Layer 12). 
\end{itemize}

\begin{figure*}[t]
    \vspace{0.85cm}
      \includegraphics[width=1\linewidth]{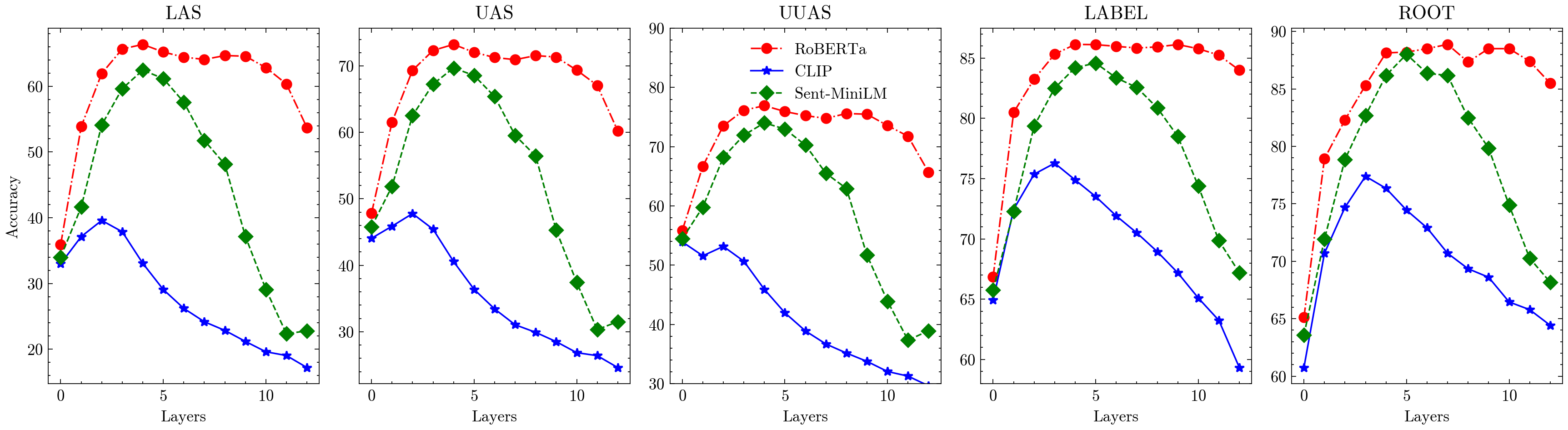}
      \caption{Comparison of the layer-wise performance between \textcolor{red}{RoBERTa}, \textcolor{blue}{CLIP} and \textcolor{ForestGreen}{Sentence-MiniLM (Sent-MiniLM)}. It can be observed that the syntactic information encoded in CLIP decreases as we go deeper into the model. Interestingly, Sent-MiniLM shows a pattern similar to RoBERTa up to the mid-layers (Layer-6) but aligns more closely with CLIP at the higher layers. This figure extends \textcolor{magenta}{Figure 3} in the main paper.}
      \label{fig:layer_probes}
  \end{figure*}


\subsection{Performance by dependency relation}

Analoguous to Tables 4 in the main paper, we provide attachment score by dependency relations in Table \ref{tab:as-zero} (Layer 0) and Table \ref{tab:as-twelf} (Layer 12).

  \begin{table}[!htb]
  \centering
  \vspace{0.2cm}
      \begin{tabular}{lrrrrr}
          \toprule
          & \multicolumn{2}{c}{CLIP} & \multicolumn{2}{c}{FLAVA} & RoBERTa \\
           & Score & $\Delta$ & Score & $\Delta$ & Score \\
          \midrule
          conj & 21.9 & 7.4 & 23.7 & 5.6 & 29.3 \\
          aux & 65.9 & 6.0 & 64.8 & 7.1 & 71.9 \\
          obj & 63.7 & 5.6 & 66.0 & 3.4 & 69.3 \\
          mark & 51.7 & 5.5 & 50.1 & 7.1 & 57.2 \\
          root & 60.7 & 4.4 & 62.0 & 3.1 & 65.1 \\
          obl & 33.0 & 4.3 & 34.5 & 2.9 & 37.4 \\
          cc & 36.2 & 3.9 & 32.3 & 7.8 & 40.1 \\
          nsubj & 51.1 & 3.8 & 50.6 & 4.3 & 55.0 \\
          cop & 44.1 & 3.7 & 40.4 & 7.5 & 47.9 \\
          advmod & 59.1 & 3.6 & 58.5 & 4.1 & 62.7 \\
          case & 28.8 & 3.4 & 27.8 & 4.4 & 32.2 \\
          det & 53.0 & 2.9 & 46.2 & 9.7 & 56.0 \\
          nmod & 42.4 & 2.9 & 42.1 & 3.2 & 45.3 \\
          amod & 42.1 & 0.7 & 44.5 & -1.8 & 42.8 \\
          compound & 39.1 & -1.2 & 37.8 & 0.1 & 37.9 \\
          \bottomrule
          \end{tabular}
          \vspace{0.1cm}
      \caption{Attachment score for the 15 most common dependency
      relations in the ground truth of the UD test split, for the zeroth (embedding) layer
      of each model. CLIP refers to the ViT-B-P32 version. $\Delta$ shows the
      performance difference to RoBERTa (base). This table Corresponds to \textcolor{magenta}{Table 4} in the main paper.}
      \label{tab:as-zero}
  \end{table}
  
  \begin{table}[!htb]\centering
  \vspace{0.2cm}
      \begin{tabular}{lrrrrr}
          \toprule
           & \multicolumn{2}{c}{CLIP} & \multicolumn{2}{c}{FLAVA} & RoBERTa \\
           & Score & $\Delta$ & Score & $\Delta$ & Score \\
          \midrule
          det & 10.7 & 64.5 & 60.0 & 15.2 & 75.2 \\
          cc & 9.2 & 54.0 & 56.8 & 6.3 & 63.2 \\
          mark & 11.9 & 51.0 & 56.5 & 6.4 & 62.9 \\
          obj & 31.5 & 44.3 & 66.1 & 9.6 & 75.8 \\
          case & 7.3 & 43.5 & 47.8 & 3.0 & 50.8 \\
          amod & 30.2 & 40.7 & 58.5 & 12.4 & 70.9 \\
          cop & 20.6 & 39.3 & 57.5 & 2.5 & 60.0 \\
          aux & 36.2 & 37.5 & 67.0 & 6.6 & 73.6 \\
          nmod & 17.3 & 36.6 & 46.3 & 7.5 & 53.9 \\
          advmod & 32.5 & 35.0 & 61.3 & 6.2 & 67.5 \\
          compound & 31.7 & 31.3 & 53.5 & 9.5 & 63.0 \\
          nsubj & 33.2 & 29.8 & 58.9 & 4.1 & 63.0 \\
          obl & 22.1 & 24.7 & 41.1 & 5.7 & 46.8 \\
          root & 64.3 & 21.2 & 82.4 & 3.1 & 85.5 \\
          conj & 21.2 & 19.6 & 38.8 & 2.0 & 40.8 \\
          \bottomrule
          \end{tabular}
          \vspace{0.1cm}
      \caption{Attachment score for the 15 most common dependency
      relations in the ground truth of the UD test split, for the twelfth layer
      of each model. CLIP refers to the ViT-B-P32 version. $\Delta$ shows the
      performance difference to RoBERTa (base). This table Corresponds to \textcolor{magenta}{Table 4} in the main paper.}
      \label{tab:as-twelf}
  \end{table}
  

  Analoguous to Tables 5 in the main paper, we provide attachment score by dependency relations in Table \ref{tab:label-zero} (Layer 0) and Table \ref{tab:label-twelf} (Layer 12). 

  \begin{table}[!htb]
  \centering
  \vspace{0.4cm}
      \begin{tabular}{lrrrrr}
          \toprule
           & \multicolumn{2}{c}{CLIP} & \multicolumn{2}{c}{FLAVA} & RoBERTa \\
           & Score & $\Delta$ & Score & $\Delta$ & Score \\
          \midrule
          compound & 37.7 & 14.4 & 39.6 & 12.4 & 52.1 \\
          root & 60.7 & 4.4 & 62.0 & 3.1 & 65.1 \\
          conj & 14.9 & 3.8 & 16.5 & 2.2 & 18.7 \\
          cop & 83.3 & 3.2 & 79.5 & 6.9 & 86.5 \\
          aux & 71.7 & 2.3 & 73.3 & 0.7 & 73.9 \\
          nmod & 40.5 & 1.4 & 42.0 & -0.1 & 41.9 \\
          advmod & 81.5 & 0.9 & 80.6 & 1.8 & 82.4 \\
          cc & 97.8 & 0.4 & 98.1 & 0.1 & 98.2 \\
          nsubj & 69.7 & 0.3 & 67.1 & 2.9 & 70.0 \\
          obl & 30.3 & 0.2 & 29.9 & 0.5 & 30.4 \\
          obj & 40.1 & -0.2 & 36.3 & 3.7 & 40.0 \\
          case & 86.8 & -0.3 & 86.8 & -0.3 & 86.5 \\
          amod & 78.0 & -0.3 & 77.1 & 0.6 & 77.7 \\
          mark & 82.5 & -0.5 & 83.1 & -1.1 & 82.0 \\
          det & 97.4 & -0.9 & 97.6 & -1.0 & 96.5 \\
          \bottomrule
          \end{tabular}
          \vspace{0.2cm}
          \caption{Labeling accuracy for the 15 most common relations in the
          ground truth of the UD test split, for the zeroth layer (embedding layer) of each model. $\Delta$ shows the performance difference to RoBERTa (base). This table Corresponds to \textcolor{magenta}{Table 5} in the main paper.}
          \label{tab:label-zero}
  \end{table}

  \begin{table}[!htb]
  \centering
  \vspace{0.4cm}
  \begin{tabular}{lrrrrr}
      \toprule
       & \multicolumn{2}{c}{CLIP} & \multicolumn{2}{c}{FLAVA} & RoBERTa \\
        & Score & $\Delta$ & Score & $\Delta$ & Score \\
      \midrule
      compound & 36.7 & 43.9 & 73.1 & 7.5 & 80.6 \\
      obj & 40.7 & 40.6 & 76.7 & 4.5 & 81.2 \\
      mark & 54.5 & 40.5 & 94.0 & 1.0 & 95.0 \\
      conj & 36.2 & 37.6 & 72.2 & 1.6 & 73.8 \\
      nsubj & 60.1 & 32.6 & 90.3 & 2.3 & 92.7 \\
      cop & 61.6 & 32.4 & 91.8 & 2.1 & 94.0 \\
      aux & 65.4 & 31.7 & 96.2 & 0.9 & 97.1 \\
      obl & 31.3 & 30.8 & 62.5 & -0.3 & 62.1 \\
      nmod & 33.8 & 30.4 & 63.3 & 0.9 & 64.2 \\
      advmod & 63.8 & 26.3 & 89.6 & 0.5 & 90.1 \\
      case & 72.8 & 24.2 & 95.6 & 1.3 & 97.0 \\
      root & 64.3 & 21.2 & 82.4 & 3.1 & 85.5 \\
      amod & 68.8 & 19.1 & 87.2 & 0.7 & 87.9 \\
      cc & 77.5 & 18.5 & 96.8 & -0.8 & 96.0 \\
      det & 85.6 & 13.2 & 99.1 & -0.3 & 98.8 \\
      \bottomrule
  \end{tabular}
  \vspace{0.2cm}
  \caption{Labeling accuracy for the 15 most common relations in the
          ground truth of the UD test split, for the 12th layer (embedding layer) of each model. $\Delta$ shows the performance difference to RoBERTa (base). This table Corresponds to \textcolor{magenta}{Table 5} in the main paper.}
      \label{tab:label-twelf}
  \end{table}

\begin{figure}[!htb]
    \begin{center}
    \begin{tabular}{rlrl}
    \includegraphics[width=.23\columnwidth]{img/stablediffusion-35-turbo/a-cat-chases-a-dog-1-real.png} &
    \includegraphics[width=.23\columnwidth]{img/stablediffusion-35-turbo/a-cat-chases-a-dog-2-real.png} &
    \includegraphics[width=.23\columnwidth]{img/stablediffusion-35-turbo/a-cat-chases-a-dog-3-real.png} & 
    \includegraphics[width=.23\columnwidth]{img/stablediffusion-35-turbo/a-cat-chases-a-dog-4-real.png} \\
    \multicolumn{4}{c}{(a) A cat chases a dog} \\
    & & & \\
    \includegraphics[width=.23\columnwidth]{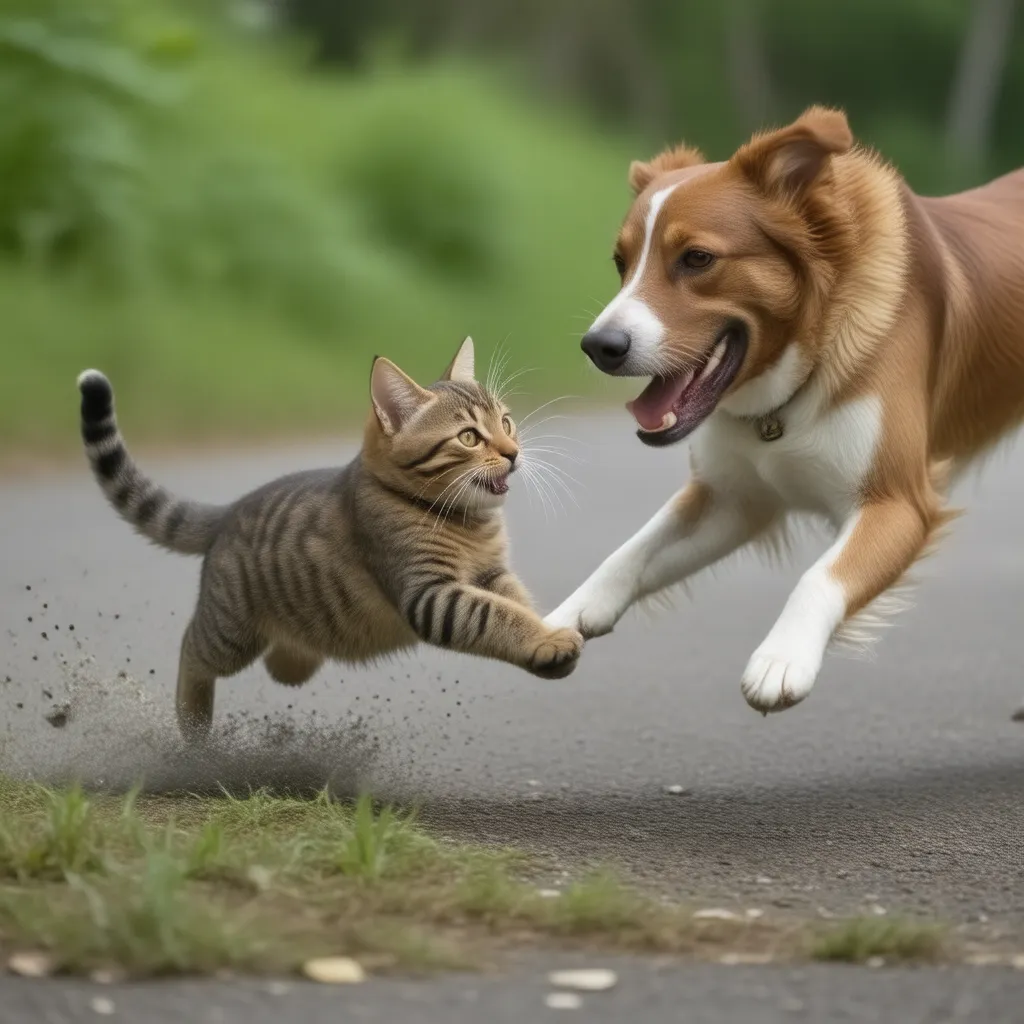} &
    \includegraphics[width=.23\columnwidth]{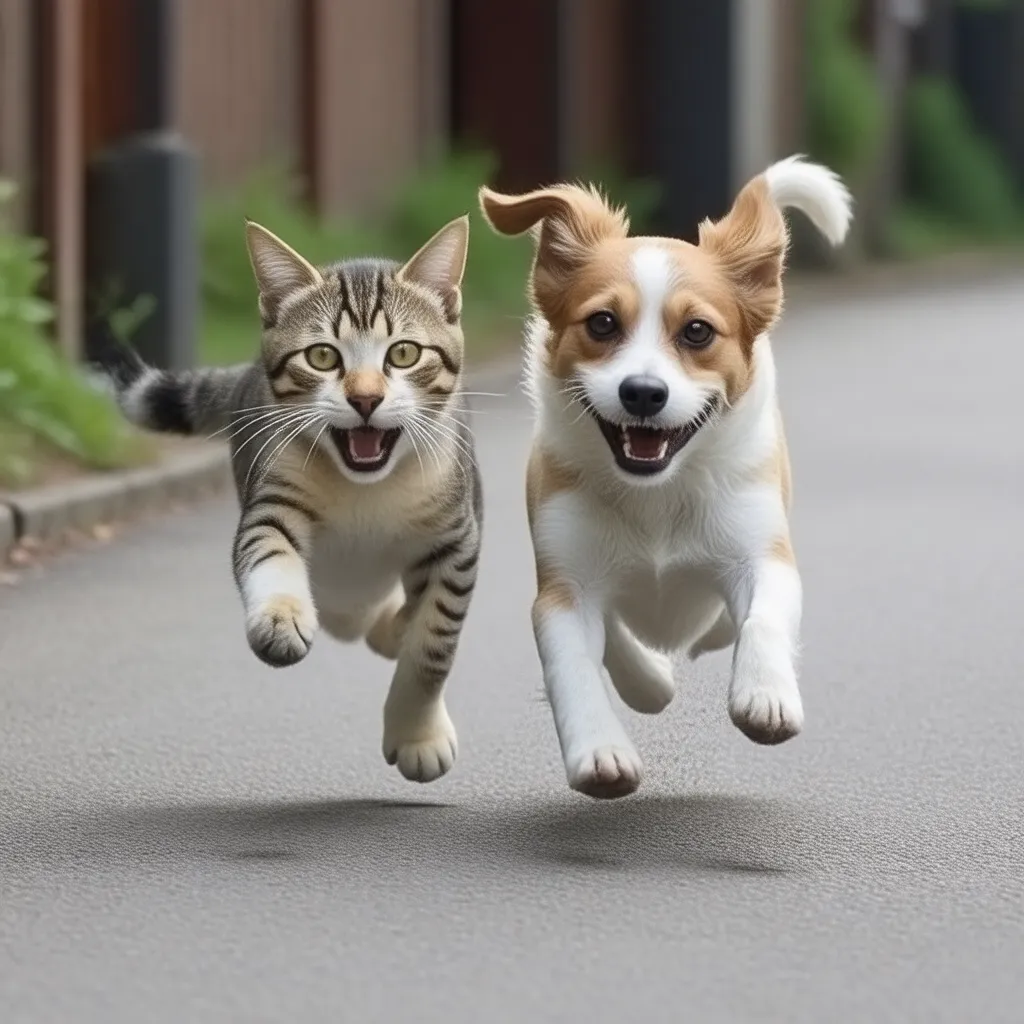} &
    \includegraphics[width=.23\columnwidth]{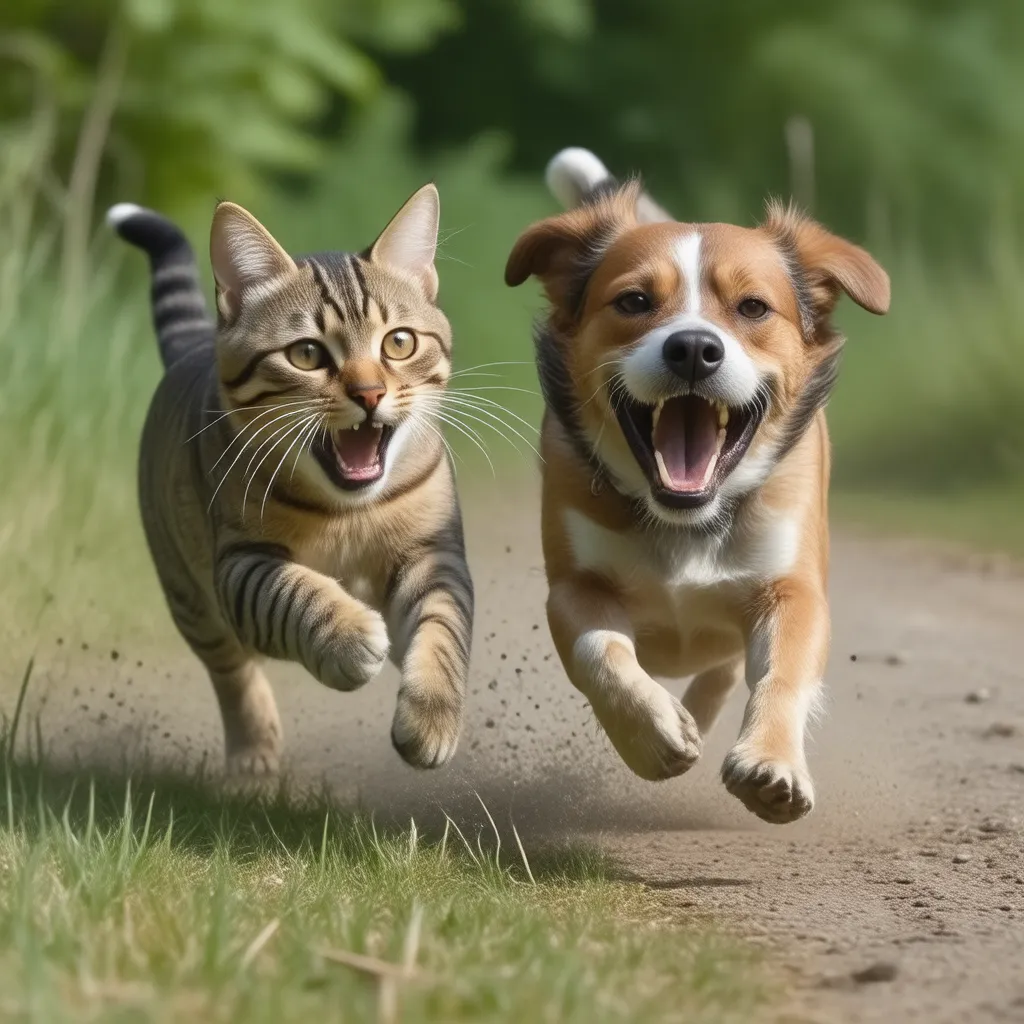} &
    \includegraphics[width=.23\columnwidth]{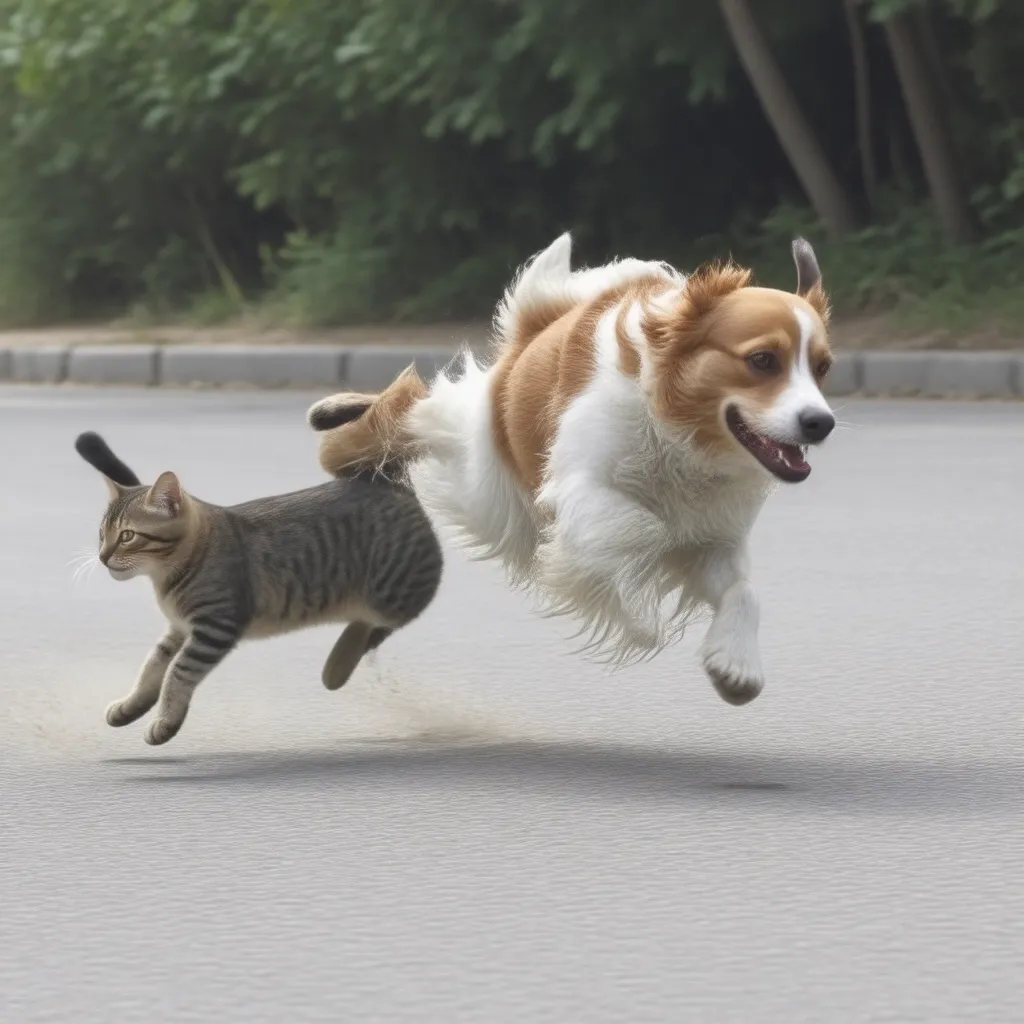} \\
    \multicolumn{4}{c}{(b) A dog chases a cat} \\
    & & & \\
    \includegraphics[width=.23\columnwidth]{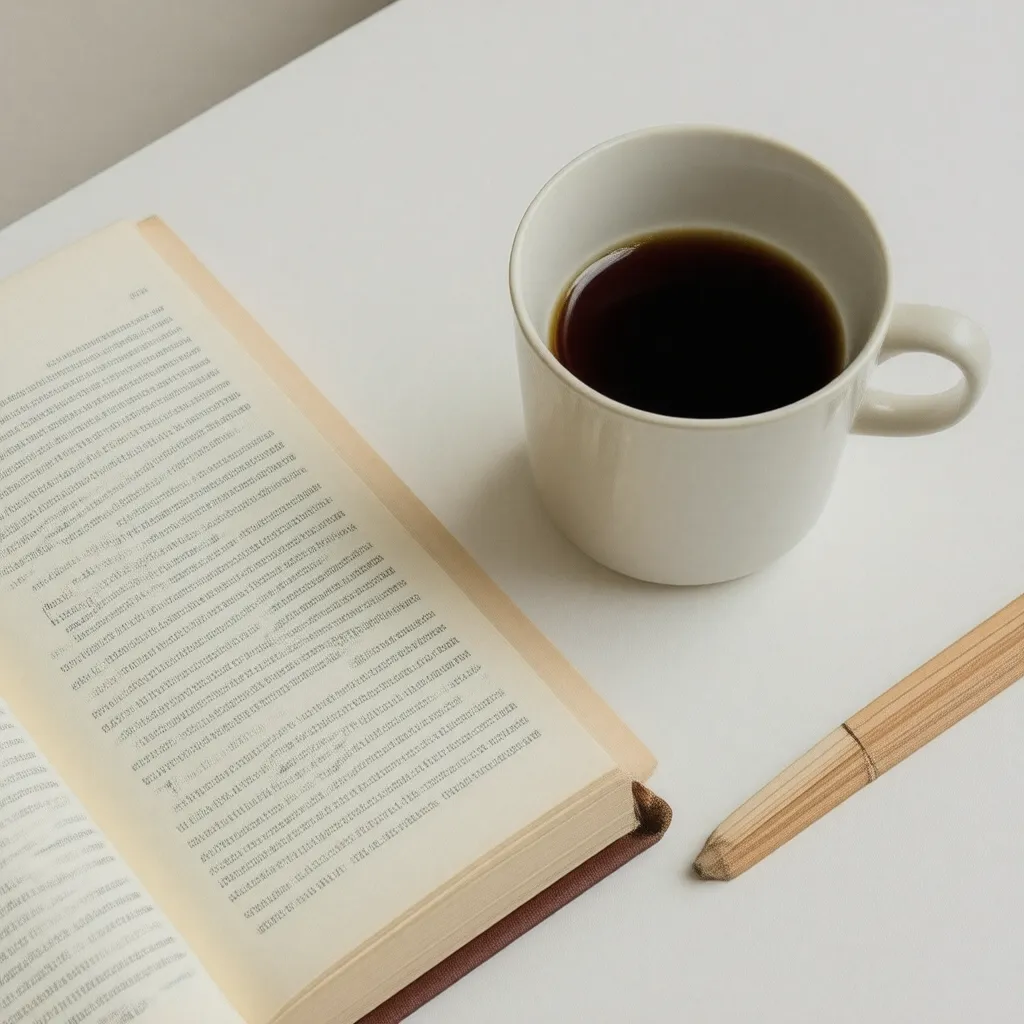} &
    \includegraphics[width=.23\columnwidth]{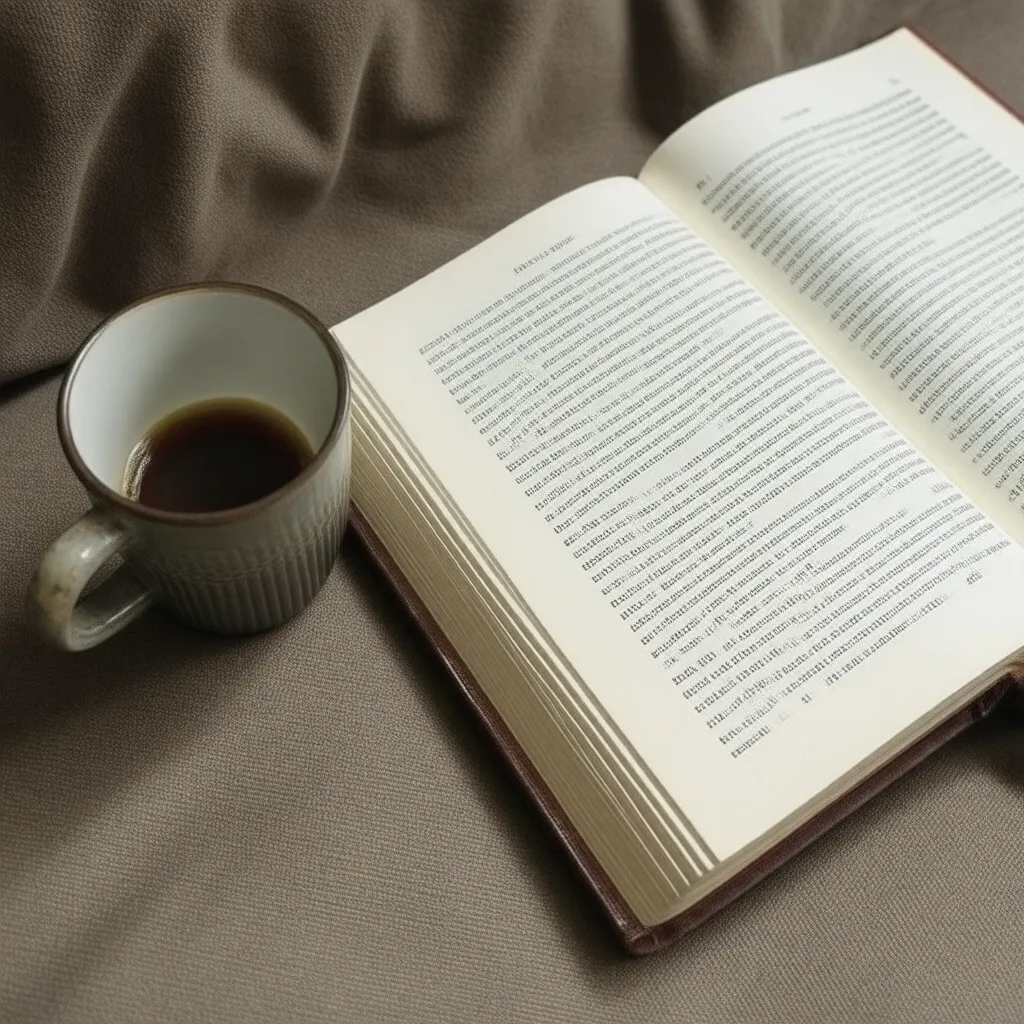} &
    \includegraphics[width=.23\columnwidth]{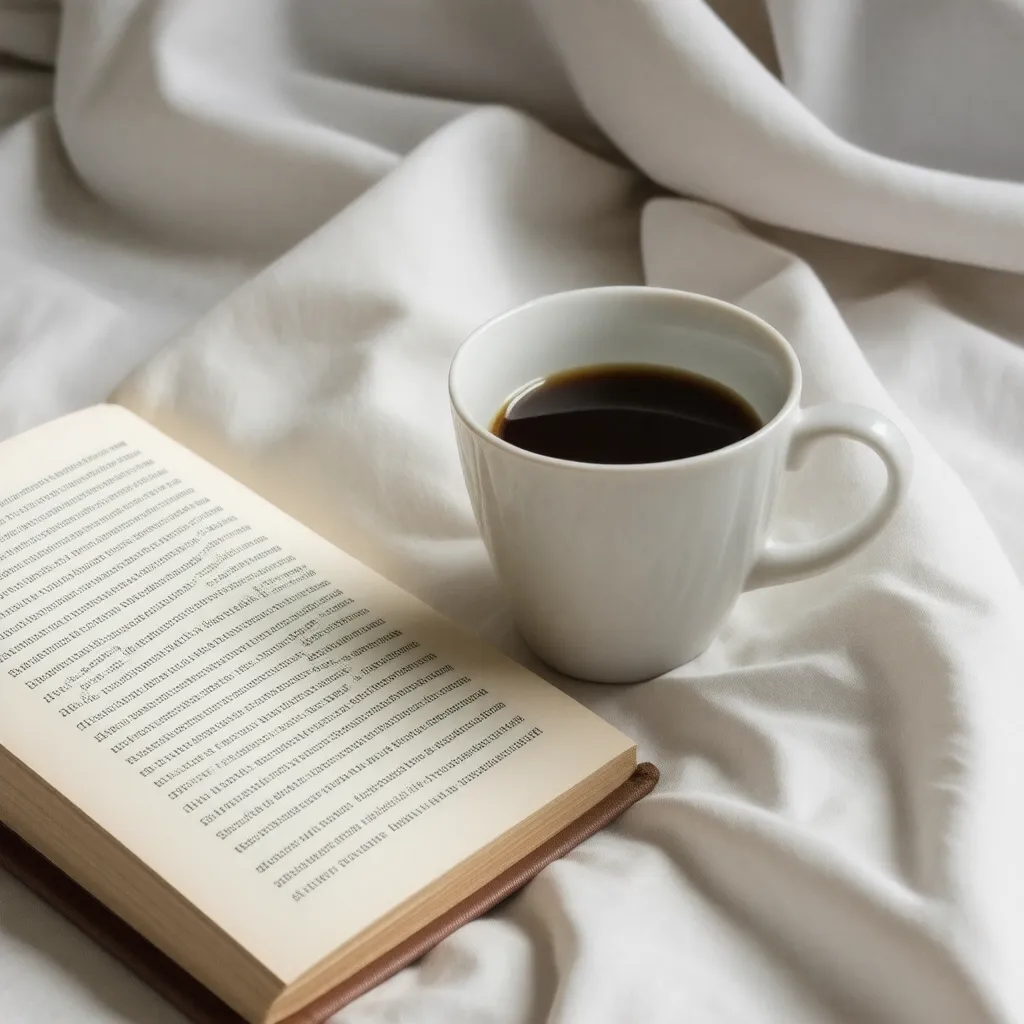} &
    \includegraphics[width=.23\columnwidth]{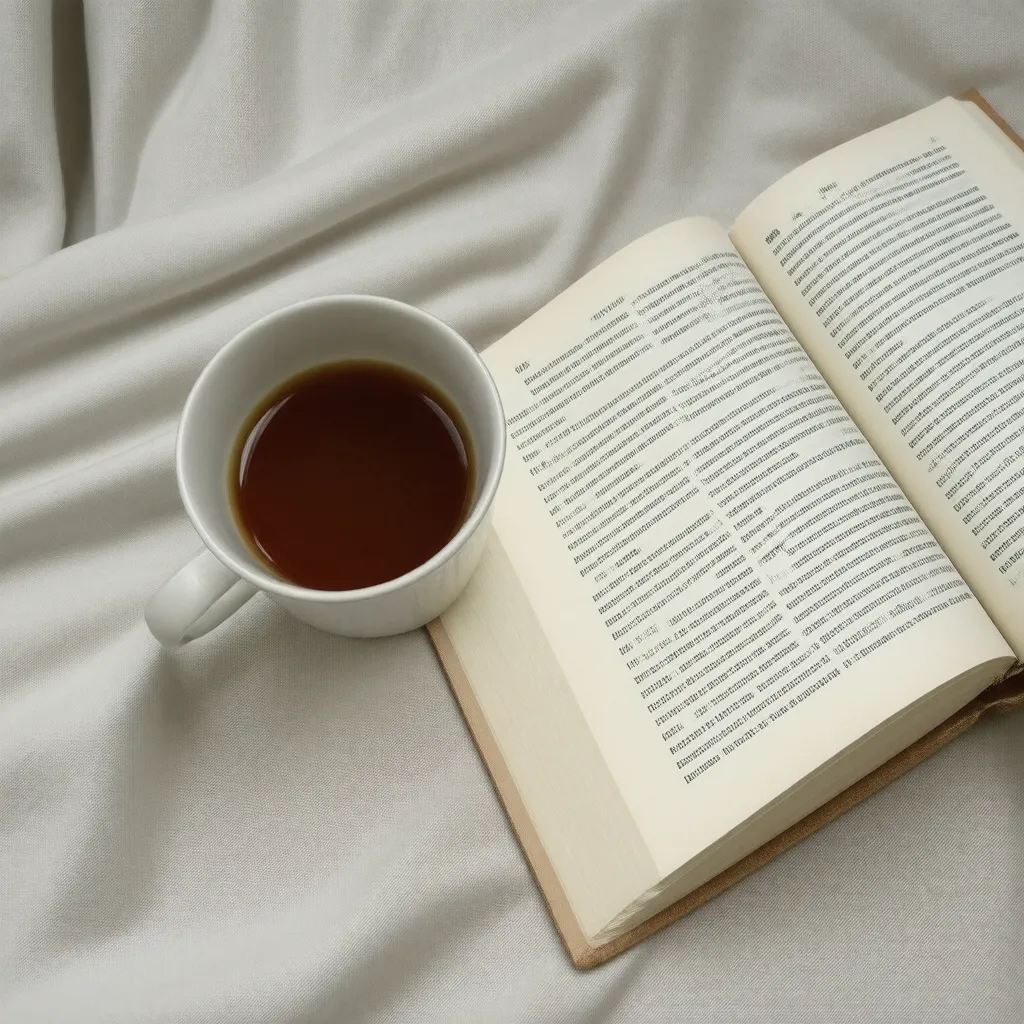} \\
    \multicolumn{4}{c}{(c) A cup to the left of a book} \\
    & & & \\
    \includegraphics[width=.23\columnwidth]{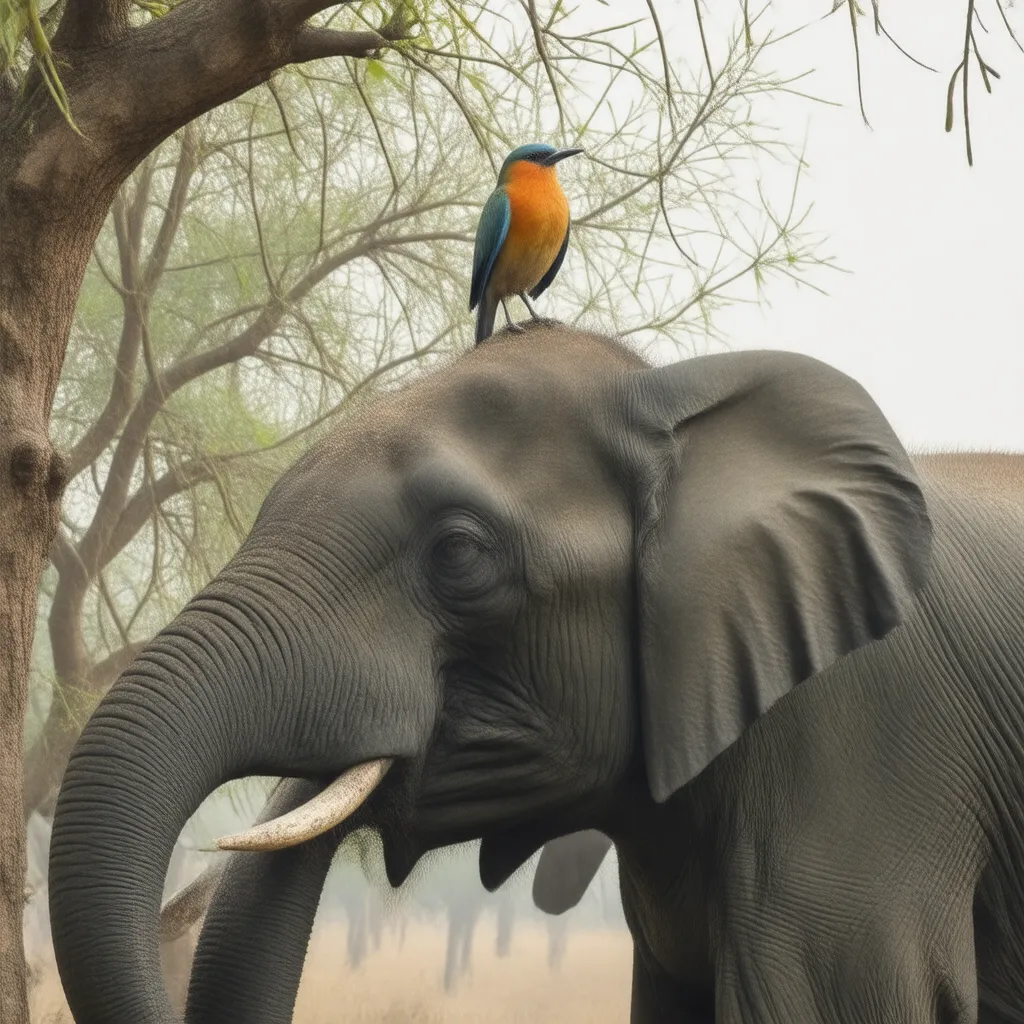} &
    \includegraphics[width=.23\columnwidth]{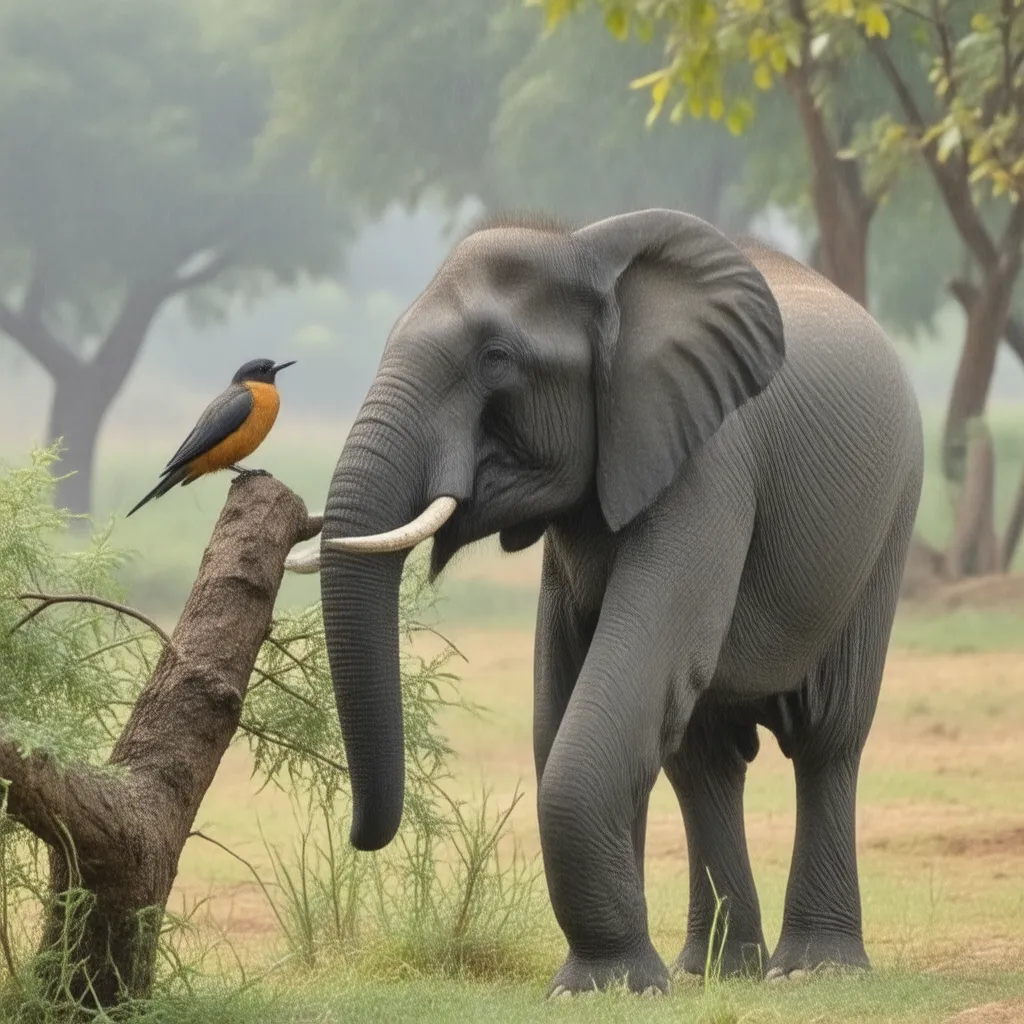} &
    \includegraphics[width=.23\columnwidth]{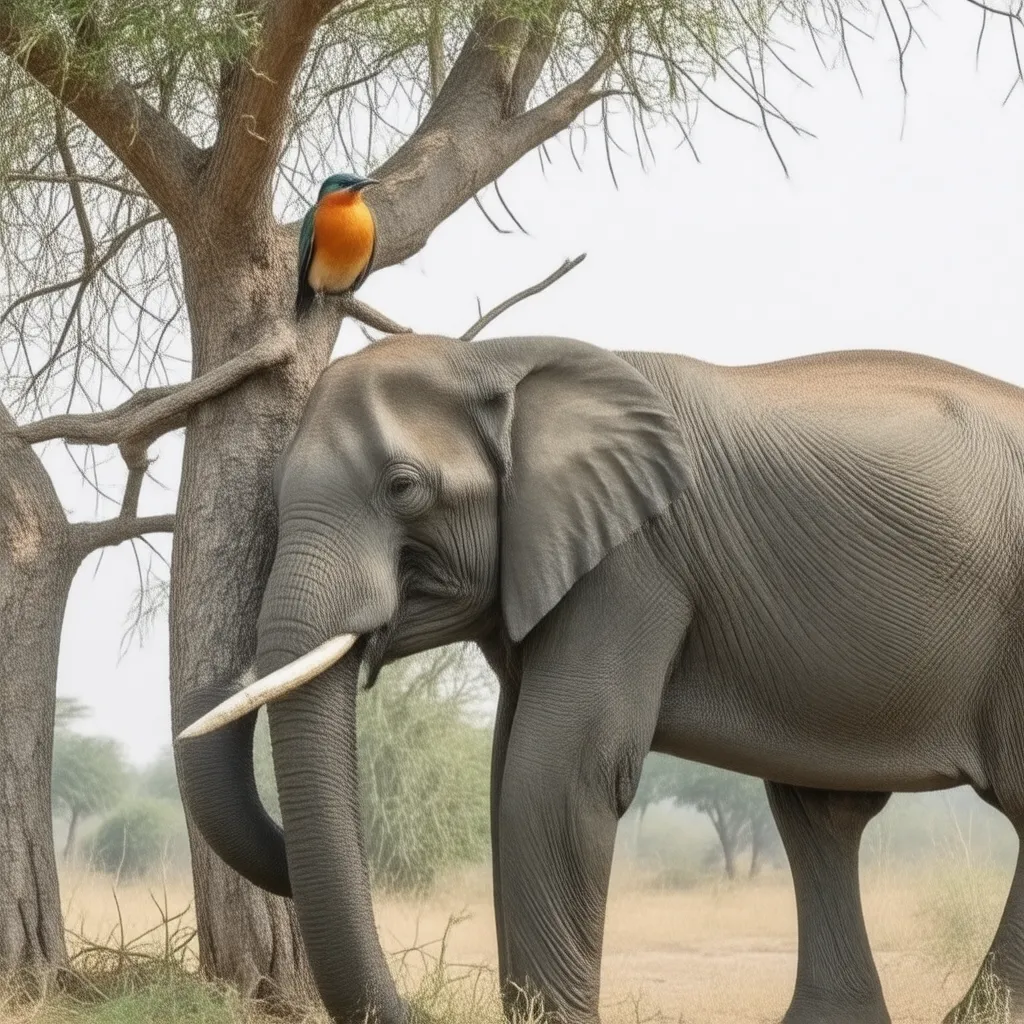} &
    \includegraphics[width=.23\columnwidth]{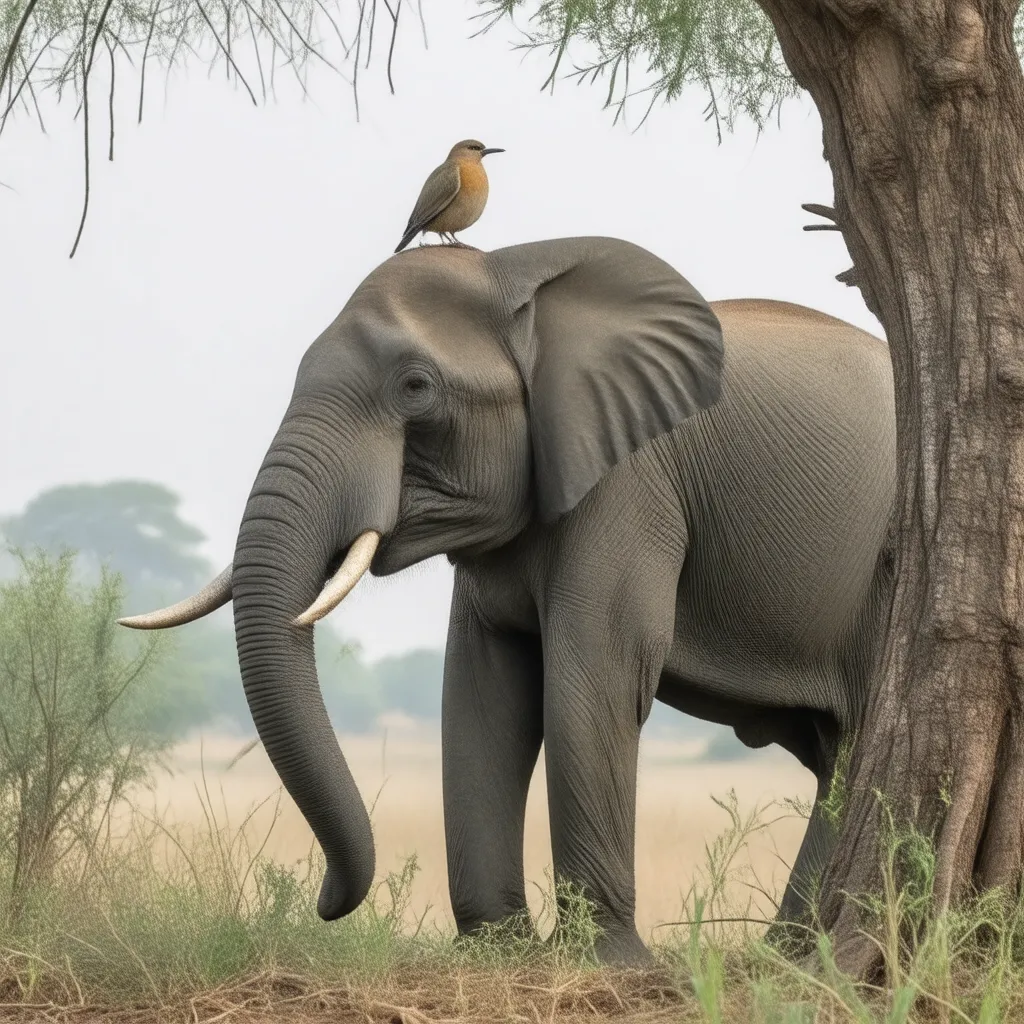} \\
    \multicolumn{4}{c}{(d) A bird sitting on a tree behind an elephant} \\
    \end{tabular}
    \end{center}
    \caption{Examples of additional images sampled from the StableDiffusion 3.5 model for different input prompts. Multiple images are generated for the same input prompt. From these images, it can be observed that the text encoders of the model identify the main components or objects in the text, such as a dog, cat, book, cup, bird, or elephant. However, they fail to encode the structure of the input text, resulting in images that reflect a random ordering of these words. This Figure is an extension of \textcolor{magenta}{Figure 1} in the main paper.}
    \label{fig:intro-fig}
\end{figure}

\section{Example Images Generated by StableDiffusion-3.5 Model}

Figure \ref{fig:intro-fig} shows the images generated by StableDiffusion-3.5 for different text inputs. StableDiffusion-3.5 uses  uses CLIP-L~\cite{radford2021learning}, CLIP-bigG~\cite{radford2021learning} and T5-XXl~\cite{t5-2020} models to encode the input prompt. We provide below four simple text inputs to the model to understand the effect of syntax understanding on image generation. \\
(a) A cat chases a dog \\
(b) A dog chases a cat \\
(c) A cup to the left of a book \\
(d) A bird sitting on a tree behind a elephant \\ 

For all four text inputs (see Figure \ref{fig:intro-fig}), the model faithfully generates the main objects—for example, the dog and cat in (a) and (b), the book and cup in (c), and the bird, tree, and elephant in (d). However, for the same text input, the model often generates images with varying arrangements of objects that do not consistently align with the input text.For example, for the input text (d) "A bird sitting on a tree behind an elephant", we observe images where the bird is sitting on the elephant with the tree behind in one instance, and in another, the bird is sitting on a tree with the elephant behind.
This indicates that while the text encoders have knowledge and understanding of the objects, they do not effectively encode the structure (syntax) of the input text.

Previous works~\cite{yuksekgonul2023and, wang2023can, kumar2024vision, dumpala2024sugarcrepe++} have shown the difficulty of vision-language models (VLMs) in encoding compositionality and semantics. However, none of these works analyzed the syntactic learning abilities of the text encoders of VLMs. In this work, we showed how the syntactic learning abilities depend on the pre-training objectives used for training VLMs. We also showed how the syntactic information is distributed across the layers of the network. We expect our research to enable future research to train better VLMs which can encode syntactic inforation better.

\end{document}